\documentclass[sn-mathphys, iicol]{sn-jnl}

\usepackage{graphicx}
\usepackage{multirow}
\usepackage{amsmath,amssymb,amsfonts}
\usepackage{amsthm}
\usepackage{mathrsfs}
\usepackage[title]{appendix}
\usepackage{xcolor}
\usepackage{colortbl}
\usepackage{textcomp}
\usepackage{manyfoot}
\usepackage{color}
\usepackage{booktabs}
\usepackage{algorithm}
\usepackage{algorithmicx}
\usepackage{algpseudocode}
\usepackage{listings}
\usepackage{tabularx}
\usepackage{makecell}
\usepackage{balance}
\usepackage{times}
\usepackage{epsfig}
\usepackage{tikz}
\usepackage{comment}
\usepackage{bbding}
\usepackage{newfloat}
\usepackage{array}
\usepackage{stfloats}
\usepackage{url}
\usepackage{verbatim}

\theoremstyle{thmstyleone}%
\theoremstyle{thmstyletwo}%

\theoremstyle{thmstylethree}%

\begin{document}

\title[Article Title]{On the Evaluation and Refinement of Vision-Language Instruction Tuning Datasets}

\author[1]{\fnm{Ning} \sur{Liao}}\email{liaoning@sjtu.edu.cn}

\author[1]{\fnm{Shaofeng} \sur{Zhang}}\email{sherrylone@sjtu.edu.cn}

\author[1]{\fnm{Renqiu} \sur{Xia}}\email{xiarenqiu@sjtu.edu.cn}

\author[3]{\fnm{Min} \sur{Cao}}\email{caomin0719@126.com}

\author[2]{\fnm{Yu} \sur{Qiao}}\email{qiaoyu@pjlab.org.cn}

\author*[1]{\fnm{Junchi} \sur{Yan}}\email{yanjunchi@sjtu.edu.cn}

\affil[1]{\orgname{Shanghai Jiao Tong University}, \orgaddress{\city{Shanghai}, \postcode{200240}, \country{China}}}

\affil[2]{\orgdiv{Shanghai Artificial Intelligence Laboratory}, \orgaddress{\city{Shanghai}, \postcode{200030}, \country{China}}}

\affil[3]{\orgname{Soochow University}, \orgaddress{\city{Suzhou}, \postcode{215006}, \country{China}}}

\abstract{
There is an emerging line of research on multimodal instruction tuning, and a line of benchmarks has been proposed for evaluating these models recently. Instead of evaluating the models directly, in this paper, we try to evaluate the Vision-Language Instruction-Tuning (VLIT) datasets. Also, we seek the way of building a dataset for developing an all-powerful VLIT model, which we believe could also be of utility for establishing a grounded protocol for benchmarking VLIT models. For effective evaluation of VLIT datasets that remains an open question, we propose a \textit{tune-cross-evaluation} paradigm: tuning on one dataset and evaluating on the others in turn. For each single tune-evaluation experiment set, we define the Meta Quality (MQ) as the mean score obtained by a set of caption metrics including BLEU, METEOR, and ROUGE-L to quantify the quality of a certain dataset or a sample. On this basis, to evaluate the comprehensiveness of a dataset, we develop the Dataset Quality (DQ) covering all tune-evaluation sets. To lay the foundation for building a comprehensive dataset and developing an all-powerful model for practical applications, we define the Sample Quality (SQ) to quantify the all-sided quality of each sample. Extensive experiments validate the rationality of the proposed evaluation paradigm. Based on the holistic evaluation, we build a new dataset, REVO-LION (REfining VisiOn-Language InstructiOn tuNing), by collecting samples with higher SQ from each dataset. Remarkably, even with only half of the complete data, the model trained on REVO-LION can achieve the performance comparable to simply adding all VLIT datasets up. Furthermore, REVO-LION not only facilitates the development of a powerful model but also incorporates an evaluation set, which is designed to serve as a convenient benchmark for future research in the field.
}


\maketitle

\section{Introduction}
The large-scale multimodal model GPT-4~\citep{openai2023gpt4} has recently exhibited strong power in generating desired answers from given images and instructions. Inspired by its remarkable success, various multimodal instruction tuning models~\citep{dai2023instructblip, chen2023visual, li2023otter, luo2023cheap} have been proposed towards different aspects of Vision-Language (VL) understanding, such as MiniGPT-4~\citep{zhu2023minigpt} for detailed description and LLaVAR~\citep{zhang2023llavar} for text-rich image understanding. With the rapid development of Vision-Language Instruction-Tuning (VLIT), evaluating these models becomes increasingly important, for which several benchmarks~\citep{yin2023lamm, xu2023lvlm, liu2023mmbench} are very recently released.
\begin{figure*}[tb!]
  \centering
  \includegraphics[width=16cm]{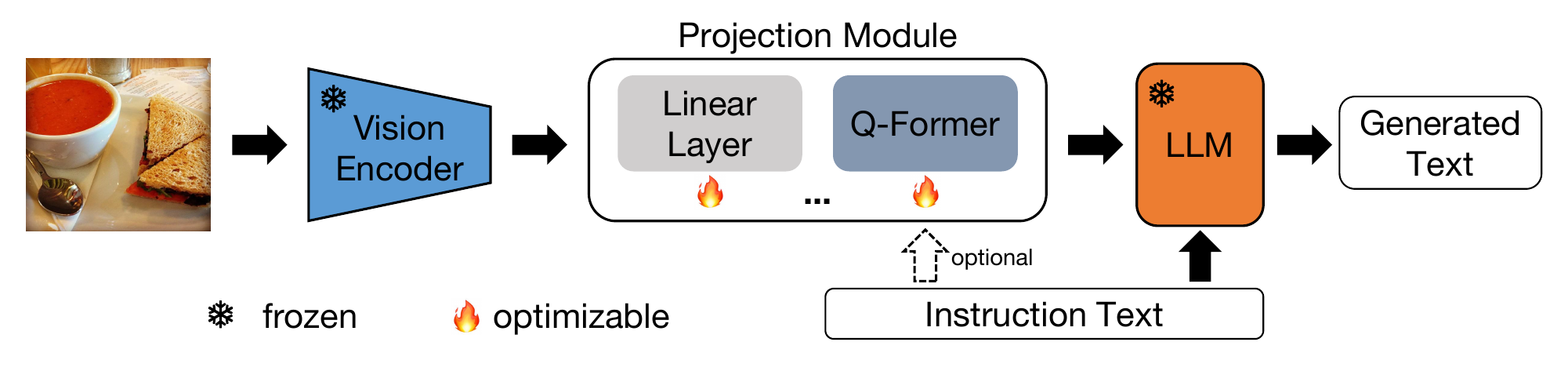}
  \caption{The popular architecture in current vision-language instruction tuning methods~\citep{dai2023instructblip, liu2023visual}. Extracting the visual feature by a frozen image encoder, transferring the visual feature into the language space using an optimizable projection module, and generating text output via a frozen Large Language Model (LLM).}
  \label{fig:VLIT_Architecture}
\end{figure*}

Different from these existing benchmarks that concentrate on evaluating VLIT models directly, our goal is one step back: \textbf{evaluating VLIT datasets}. The motivation comes from the insights into current VLIT models, including two similarities and one difference. \textit{The first similarity is the model architecture} as shown in Fig.~\ref{fig:VLIT_Architecture}. The image feature is firstly extracted by a frozen vision encoder~\citep{fang2023eva}. Then, a learnable projection module, which can be simply designed either as the linear layer in LLaVA~\citep{liu2023visual} or a more sophisticated one like Q-Former in InstructBLIP~\citep{dai2023instructblip}, transforms the image feature to the text space. Finally, by feeding the transformed image feature and instruction text into the frozen Large Language Models (LLMs)~\citep{touvron2023llama, vicuna2023}, the instruction-following answer is generated. \textit{The second similarity is the two-stage learning scheme}. During training, common large-scale image-text pairs~\citep{ordonez2011im2text, schuhmann2021laion, sharma2018conceptual} are leveraged for the cross-modal feature alignment in the first stage. Then, the customized high-quality instruction data is used to train the VLIT model to generate coherent and desired output in the second stage. \textit{The difference is exactly the high-quality instruction data targeting at different aspects of VL understanding}, as concluded in Table~\ref{tab:VLIT_data_statistic}. To be more consistent with LLMs, the annotations in these datasets are almost generated or augmented by GPT. It follows that curating proper instruction tuning datasets is essential in VLIT, which motivates us to evaluate VLIT datasets and look into their quality.

However, there exist limitations when using current benchmarks for evaluation. For example, in benchmarks~\citep{xu2023lvlm, liu2023mmbench}, the style of annotations is quite different from the style of the open-ended texts generated by LLMs, causing possible bias for assessment. Besides, human voting~\citep{xu2023lvlm} and ChatGPT/GPT-4~\citep{openai2023gpt4} are leveraged for performance evaluation. While the former is labor-intensive and liable to cause subjective evaluation, and the latter is inconvenient and unstable for widespread use because of the API availability and the changeable output. Additionally, it is worth noting that although evaluations utilizing GPT models have demonstrated the highest agreement with human evaluations~\citep{bitton2023visit}, both GPT models and human evaluation are not well-suited for large-scale evaluation scenarios due to practical considerations.

To conduct a comprehensive analysis of VLIT datasets, we introduce a pioneering \textit{tune-cross-evaluation} paradigm, as shown in Fig.~\ref{fig:DQ_SQ}, based on the common architecture in Fig.~\ref{fig:VLIT_Architecture}. This paradigm allows us to thoroughly assess the datasets. The fundamental concept is that each dataset serves a dual purpose: it can be utilized for model development and also function as a benchmark for the specific aspect it was designed to address. Our evaluation paradigm benefits from annotations consistent with LLMs, enabling us to define the Meta Quality (MQ) as the average score measured by caption metrics, including BLEU, METEOR, and ROUGE-L. This model-free and human-free evaluation strategy, utilizing MQ to measure performance in each tune-evaluation experiment set, offers greater convenience and stability than GPT-involved scoring and a more objective assessment than human voting. Building upon the proposed MQ, we devise the concepts of Dataset Quality (DQ) and Sample Quality (SQ) to measure the overall capability of each dataset and sample, combining all tune-evaluation sets. 

Taking a step further, the other goal in this paper is \textbf{refining VLIT datasets} according to the holistic evaluation on a set of VLIT datasets. On one hand, existing VLIT models are only equipped with one or several abilities in VL understanding, which leads to unsatisfying performance in comprehensive evaluations. On the other hand, existing benchmarks build evaluation datasets by collecting datasets from different tasks~\citep{krizhevsky2009learning, lu2022learn} with annotations inconsistent with the open-ended generated texts~\citep{xu2023lvlm, yin2023lamm}, which causes inaccurate evaluation. As a result, a dataset encompassing multiple VLIT capabilities is critical for developing an all-powerful model and building an unbiased benchmark in a convenient way.

To this end, we build the so-called REVO-LION dataset by REfining VisiOn-Language InstructiOn tuNing datasets, which is composed of samples with higher SQ from each dataset as listed in Table~\ref{tab:VLIT_data_statistic}. As a compact subset of the original datasets, REVO-LION is shown empirically to be more sample-efficient than simply merging the raw datasets together, which validates the effectiveness of the proposed SQ and the refinement strategy. We make following contributions:
\begin{enumerate}[$\cdot$]
\item[$\bullet$] We propose a \textit{tune-cross-evaluation} paradigm on VLIT datasets. To our best knowledge, this is the first holistic analysis on VLIT datasets.
\item[$\bullet$] We define a model-free and human-free evaluation metric known as the Meta Quality (MQ), as the mean score measured by BLEU, METEOR, and ROUGE-L. Based on MQ, Dataset Quality (DQ) and Sample Quality (SQ) are devised to quantify the comprehensive quality of each dataset and sample in VLIT, respectively.
\item[$\bullet$] We collect and release a comprehensive dataset called REVO-LION, by refining public mainstream VLIT datasets. REVO-LION consists of a training set for developing a highly capable VLIT model and an evaluation set that serves as an effective benchmark. Extensive experiments show that in addition to expanding the data scale, collecting high-quality data based on rational and effective evaluation is crucial for the advancement of VLIT models. 
\end{enumerate}

\section{Related Work}
\subsection{Vision-Language Instruction Tuning}
\label{sec:VLIT_related}
With the success of ChatGPT and InstructGPT~\citep{ouyang2022training} in solving tasks aligned with human instructions, subsequent Large Language Models (LLMs)~\citep{peng2023instruction, ding2023enhancing, zhou2023lima, du2022glm} have been further devised by fine-tuning open-source LLMs such as LLaMA~\citep{touvron2023llama} and GLM~\citep{zeng2022glm} using instruction data in the last two years. For example, Vicuna-13B~\citep{vicuna2023} is supervised fine-tuned from LLaMA-13B~\citep{touvron2023llama} using 70K shared conversations with ChatGPT from ShareGPT.com; Alpaca-7B~\citep{alpaca} is fine-tuned from LLaMA-7B~\citep{touvron2023llama} on 52K instruction-following demonstrations generated by Self-Instruct~\citep{wang2022self}.

Standing on the shoulder of LLMs, many Vision-Language Instruction Tuning (VLIT) models~\citep{su2023pandagpt, ye2023mplug, luo2023cheap, li2023otter, aiello2023jointly} have been proposed within a year. These models are similarly constructed by using a projection module to connect the pre-trained vision model for visual perception and the language model for text generation. The projection module is firstly trained on common image-text pairs for VL alignment, then on high-quality data for instruction tuning. One of the most impactful methods is InstructBLIP~\citep{dai2023instructblip}, which is built upon the VL alignment achieved by the Q-Former in BLIP2~\citep{li2023blip}. After collecting and transforming 28 datasets from 11 tasks into instruction format, InstructBLIP~\citep{dai2023instructblip} takes the instruction as a guidance of Q-Former to extract instruction-aware visual features for further tuning. Similar to InstructBLIP, MiniGPT-4~\citep{zhu2023minigpt} is firstly pre-trained on large-scale datasets~\citep{ordonez2011im2text, schuhmann2021laion} for VL alignment, then curates around 3500 high-quality instruction data, with the assistance of ChatGPT and Vicuna~\citep{vicuna2023} targeting at comprehensive image description, for instruction tuning in the second stage. LRV~\citep{liu2023aligning} constructs a dataset including both positive and negative instructions for robust tuning with mitigated hallucination issues based on MiniGPT-4. Simpler than MiniGPT-4, LLaVA~\citep{liu2023visual} adopts a linear layer to bridge the gap between visual and language space in the first stage using 595K image-text pairs filtered from CC3M. Then, by using ChatGPT and GPT-4, 158K instruction samples including conversations, detailed descriptions, and complex reasoning are collected in LLaVA for instruction tuning in the second stage. Similar to LLaVA~\citep{liu2023visual}, DetGPT~\citep{pi2023detgpt} collects around 30K query-answer pairs towards reasoning-based object detection for instruction tuning in the second stage, LLaVAR~\citep{zhang2023llavar} enhances the text-rich image understanding ability by collecting 16K text-rich image data, Macaw~\citep{lyu2023macaw} builds a dataset consisting of 69K instances for human-style text generation.

To make a brief summary, existing VLIT models mostly share the similar model architecture and the two-stage learning scheme. The major difference lies in the high-quality instruction data used in the second stage. Beyond current VLIT models targeting at certain aspects, as concluded in Table~\ref{tab:VLIT_data_statistic}, collecting a comprehensive dataset lies the foundation for developing an all-powerful VLIT model.
\begin{table*}[tb!]
     \small
	\centering
	\caption{Popular vision-language instruction tuning datasets on current VLIT methods. These datasets are used to build the proposed REVO-LION in this paper.}
  \label{tab:VLIT_data_statistic}
  \begin{tabular}{ c@{} | c@{} | c@{}  }
    \toprule
    Datasets & Size &  Purpose  \\
    \midrule
    DetGPT~\citep{pi2023detgpt} & \,\, 50K images and around 30K query-answer pairs \,\, & \,\, Reasoning-based object detection. \,\,\\
    \midrule
    LAMM~\citep{yin2023lamm} & 186K image-language instruction-response pairs & \makecell{Daily conversation,\\factual knowledge reasoning,\\detailed description,\\visual task dialogue.}\\
    \midrule
    \,\,\,LLaVAR~\citep{zhang2023llavar}\,\,\, & 16K high-quality instruction following data. & Text-rich image understanding\\
    \midrule
    LLaVA~\citep{liu2023visual} & \makecell{58K in conversations,\\23K in detailed description,\\77K in complex reasoning.} & \makecell{Conversations,\\detailed description,\\complex reasoning.}\\
    \midrule
    Macaw~\citep{lyu2023macaw} & 69K image instances. & Human-written style text generation.\\
    \midrule
    MiniGPT-4~\citep{zhu2023minigpt} & Around 3.5K image-text pairs. & Comprehensive image description. \\
    \midrule
    LRV~\citep{liu2023aligning} & Around 120K instances. & \makecell{Robust visual instruction with\\ mitigated hallucination issue.}\\
    \bottomrule
  \end{tabular}
\end{table*}

\subsection{VLIT Benchmarks}
With the rapid development of VLIT models, how to comprehensively and effectively evaluate these models becomes a concurrent significant problem. To this end, several benchmarks~\citep{zeng2023matters, yu2023mm, bitton2023visit} have been proposed in the last few months. The pioneering benchmark is the LVLM-eHub~\citep{xu2023lvlm}, which evaluates VLIT models by quantifying the performance and human voting in the online arena platform. In performance quantification, they utilize 47 standard benchmarks covering 6 capabilities for evaluation, and find that image caption metrics are ineffective because the style of open-ended generated texts differs from that of annotations in the benchmarks. It is reasonable because the annotations in these benchmarks are rough, simple and outdated in the context of LLMs. Immediately after LVLM-eHub, LAMM~\citep{yin2023lamm} is proposed for evaluation on 9 common image tasks by collecting 11 datasets. Except for task-specific metrics, LAMM adopts GPT as a judgment for performance evaluation. However, MME~\citep{fu2023mme} argues that human voting and GPT scoring bring problems of subjectivity and inaccuracy. For this, MME exams perception and cognition abilities covering 14 subtasks by manually constructing instruction-answer pairs and leading the tested models to answer ``yes" or ``no", which is designed for objective and accurate quantitative statistics. Nevertheless, such performance evaluation that heavily relies on generating ``yes" or ``no" is not quite reasonable, because existing VLIT models usually target at detailed tasks instead of making decisions from ``yes" or ``no" strictly. For fine-grained ability assessment, MMBench~\citep{liu2023mmbench} curates a dataset covering 20 fine-grained skills, and all instances are transformed into multi-choice problems. For robust evaluation, it employs ChatGPT for answer extraction and judgment in the proposed circular evaluation strategy, which is unable to evaluate the models directly on the generated texts, causing inaccurate assessment.

In short, there are three aspects that are not fully satisfied in existing benchmarks: 1) collecting datasets with annotations consistent with open-ended generated texts for evaluation; 2) avoiding human subjectivity in data selection and evaluation; 3) designing stable and convenient quantification metrics. We argue that it is possible to meet these conditions via our \textit{tune-cross-evaluation} paradigm on VLIT datasets with the proposed quality metrics at both dataset and sample levels. In particular, based on our deep dive in Sec.~\ref{sec:VLIT_related}, we propose shifting the focus of model evaluation, which existing benchmarks are paying great efforts on, to dataset evaluation.
\begin{figure*}[tb!]
\centering
\includegraphics[width=15.5cm]{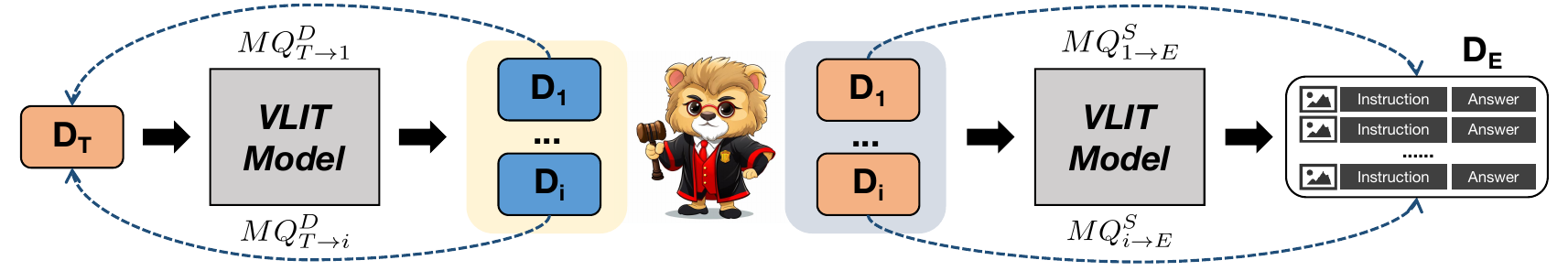}
\caption{The overall framework of the proposed \textit{tune-cross-evaluation} paradigm. \textit{Left}: The diagram of Dataset Quality (DQ) evaluation. Each dataset adopted for testing measures the quality of the tuning dataset $D_T$ on the aspect that the testing datasets are constructed towards. \textit{Right}: The diagram of Sample Quality (SQ) evaluation. Each dataset used for tuning measures how well the samples in the testing set $D_E$ match with the ability that the tuning dataset is constructed towards.}
\label{fig:DQ_SQ}
\end{figure*}

\section{Methodology}
\subsection{Tune-Cross-Evaluation Paradigm}
As shown in Fig.~\ref{fig:DQ_SQ}, we propose the \textit{tune-cross-evaluation} paradigm to evaluate VLIT datasets, which are specifically listed in Table~\ref{tab:VLIT_data_statistic}. Note that these datasets are all in English such that we do not need to handle the language bias problem, which is also not the focus of this paper. On one hand, each dataset is employed to develop a model by instruction tuning. On the other hand, because these VLIT datasets are almost annotated by leveraging GPT-4~\citep{openai2023gpt4} or ChatGPT for text generation or augmentation, each dataset also represents a standard on the aspect that the dataset is constructed towards, by which the proper annotations consistent with open-ended generated texts are accessible. Based on the VL alignment learned in the first stage by the model with the architecture in Fig.~\ref{fig:VLIT_Architecture}, at each time, we select one dataset from these datasets for instruction tuning, and the remaining datasets are used for test at this time. For example, when we use DetGPT~\citep{pi2023detgpt} for instruction tuning, the tuned model equipped with great reasoning-based object detection ability will be further tested on other datasets, and they involve testing the model's ability such as daily conversation, factual knowledge reasoning, detailed description, etc. By taking turns to cycle in this way, we finally get the comprehensive quality evaluation of each dataset and each sample. To quantify the comprehensiveness, we define the Meta Quality (MQ), Dataset Quality (DQ) and Sample Quality (SQ), and detail them in the following sections.

\subsection{Meta Quality (MQ)}
In LVLM-eHub~\citep{xu2023lvlm}, the authors show that metrics in caption tasks are ineffective for VLIT evaluation due to the style differences between the diverse open-ended generated texts and the ground-truths in the datasets curated prior to LLMs, which are outdated compared to LLMs. Benefiting from the proposed \textit{tune-cross-evaluation} paradigm, when making full use of VLIT datasets as evaluation ones, the proper annotations, which are created by GPT models to be consistent with LLMs, are available. Therefore, with mitigated style differences, to perform a model-free and human-free evaluation, in which we do not rely on other models such as GPT or human for scoring, we define the Meta Quality (MQ) as the average of scores measured by caption metrics to quantify the one-side quality of each dataset or sample within a single tune-evaluation experiment. Considering the time-consuming process in calculating sample-wise MQ if using SPICE, we use BLEU@1 (B@1), BLEU@2 (B@2), BLEU@3 (B@3), BLEU@4 (B@4), METEOR (M), and ROUGE-L (R) as the components for MQ definition. CIDEr is set as a hold-out metric in data refinement in Sec.~\ref{sec:ablate_refine}. The MQ is formulated as:
\begin{equation}
\begin{aligned}
MQ = mean(B@1+B@2+B@3+B@4+M+R).
\label{eq:MQ}
\end{aligned}
\end{equation}

The ablation of the combinations is studied in Sec.~\ref{sec:ablate_MQ}. It should be noted that the MQ can be commonly used to measure on a set of samples. When the number of samples is 1, it actually measures the sample-wise quality. For distinction, we denote the MQ measured on a dataset and a sample as $MQ^D$ and $MQ^S$, respectively.

\subsection{Dataset Quality (DQ)}
In the proposed \textit{tune-cross-evaluation} paradigm, each time we select a dataset denoted as $D_T$ from the set of datasets $S$ for instruction tuning, the remaining datasets denoted as $D_i (i \in S, i \neq T)$ are then leveraged as evaluation ones for inference, thus measuring the quality of the tuning dataset on the aspect that the evaluation datasets are constructed towards one by one, as shown on the left side of Fig.~\ref{fig:DQ_SQ}. Note that though the datasets to be evaluated in this paradigm are in different sizes, we do not explicitly separate a metric for data size. When using a dataset as an evaluation one, its size has been implicitly integrated into the measurement. A dataset with larger size is more likely to exhibit better comprehensive ability, as compared between LLaVA-Detailed description and MiniGPT-4, which both concentrate on detailed image description in Table~\ref{tab:DQ} in experiments. In a single tune-evaluation set, the one-side dataset quality is denoted as $MQ^D_{T\rightarrow i}$, in which the right arrow indicates the direction from the tuning dataset to the evaluation dataset. Specifically, we set the quality $MQ^D_{T\rightarrow T}$ that each tuning dataset exhibits on its aspect as 1, the maximum value of $MQ$. Therefore, when a dataset is set as the tuning one, its comprehensive quality measured by all capabilities in $S$ is formulated as the sum of all one-side qualities:
\begin{equation}
\begin{aligned}
DQ_T &= MQ^D_{T\rightarrow T} + \sum\limits_{i \in S, i \neq T} {{MQ^D_{T\rightarrow i}}}\\
&= 1 + \sum\limits_{i \in S, i \neq T} {{MQ^D_{T\rightarrow i}}}, T \in S.
\label{eq:DQ}
\end{aligned}
\end{equation}

By setting each dataset as the tuning one and the remaining as evaluation ones in turn, the comprehensive DQ, which measures various capabilities, for all datasets can be calculated. 

\subsection{Sample Quality (SQ)}
\label{sec:SQ}
Because the MQ can only be calculated on the inference datasets, it is hard to measure the quality of each sample in the tuning dataset when keeping the same evaluation direction in DQ, i.e., the inference datasets are regarded as standards. In contrast, when a dataset $D_E$ is set as the inference one, we hold that the model equipped with the ability of dataset $D_i (i \in S, i \neq E)$, after tuned on which, is supposed to be a standard. By this way, the $MQ^S_{i \rightarrow E}$ for each sample in $D_E$ measures how close the sample matches with the ability of the tuning dataset $D_i$, as shown on the right side of Fig.~\ref{fig:DQ_SQ}. To calculate the comprehensive quality that each sample exhibits on other aspects, other than DQ having the ability corresponding to itself, we define the SQ as a weighted sum:
\begin{equation}
\begin{aligned}
SQ_E = \sum\limits_{i \in S, i \neq E} {{ DQ_i \cdot MQ^S_{i\rightarrow E}}}.
\label{eq:SQ}
\end{aligned}
\end{equation}

We use the $DQ_i$ as the weights for objective evaluation, the higher $DQ_i$ represents a more confident evaluation when using dataset $D_i$ to tune the model, which is regarded as a standard. By setting each dataset as the inference one and the remaining as tuning ones in turn, the comprehensive SQ for each sample exhibits on other datasets can be calculated.

\subsection{REVO-LION}
To build a comprehensive dataset integrating all capabilities of the evaluated datasets, a simple yet direct way is to merge these datasets into one without more operations. As suggested in the analysis~\citep{zeng2023matters}, data quality is more significant than data quantity. Therefore, we propose to REfine VisiOn-Language InstructiOn tuNing (REVO-LION) datasets according to the proposed SQ, which measures the comprehensive quality of each sample exhibits on other datasets. To preserve all capabilities, we collect samples with higher SQ from each dataset to compose REVO-LION. Formally, we denote the portion that the number of selected samples to the number of all samples in each dataset as $P$. The lower bound of SQ in dataset $D_i (i \in S)$ corresponding to the portion $P$ is $\tau^P_i$. For each sample $\boldsymbol{x}^k_i \in D_i$, if the SQ of it $SQ^k_i$ is no lower than $\tau^P_i$, the sample is collected in REVO-LION, which is formulated as:
\begin{equation}
\begin{aligned}
S1 = \bigcup\limits_{i \in S} \boldsymbol{x}^k_i, \quad (\boldsymbol{x}^k_i \in D_i, SQ^k_i >= \tau ^P_i).
\label{eq:revolion}
\end{aligned}
\end{equation}

We denote this refinement strategy as $S1$, which is validated to be more effective than ``Random Refinement" ($S2$) and ``Gaussian Refinement" ($S3$) in Sec.~\ref{sec:ablate_refine}. After performing the data evaluation and creating REVO-LION from the datasets in Table~\ref{tab:VLIT_data_statistic}, we split it into a training set and an evaluation set. The former can serve as a common dataset for developing an all-powerful VLIT model, and the latter can serve as a convenient benchmark covering all capabilities of these datasets and equipping with ideal annotations, based on which the caption metrics can be conveniently employed for model-free and human-free evaluation.
\begin{figure*}[tb!]
\centering
\includegraphics[width=15cm]{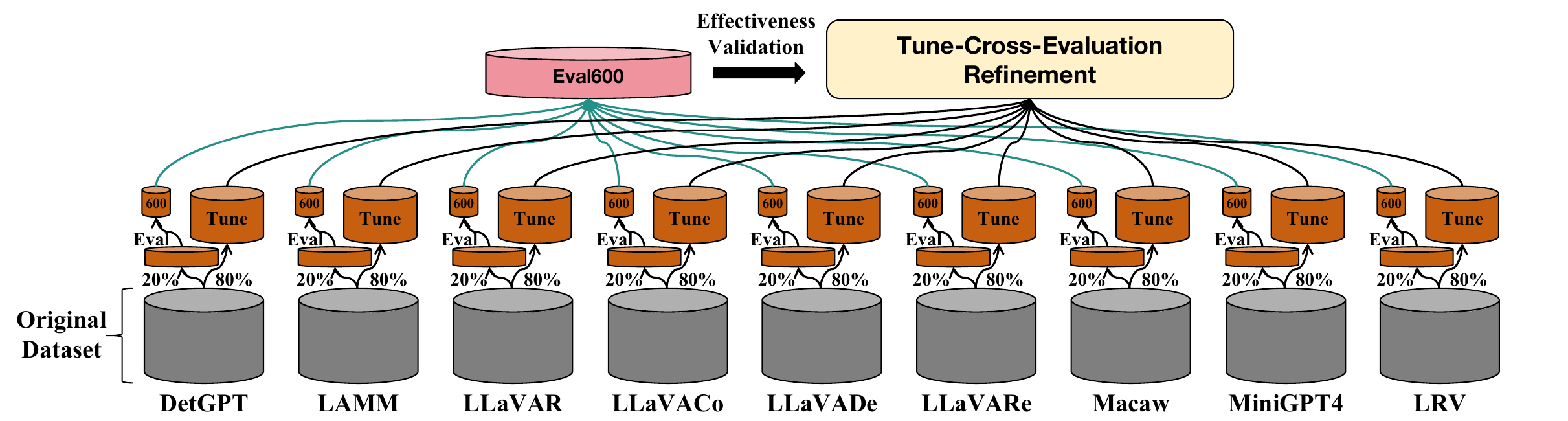}
\caption{The diagram of the data split process. It is designed to validate the effectiveness of the proposed \textit{tune-cross-evaluation} paradigm and the data refinement strategy in main experiments. Each original dataset is divided into two parts: $80\%$ samples are collected as a tuning set for data evaluation and refinement, and 600 samples from the remaining $20\%$ are collected into a balanced and comprehensive evaluation set. For robust validation, we perform such partitions twice, thus creating SPLIT1 and SPLIT2 that are used in the main experiments.} 
\label{fig:Effect_Val}
\end{figure*}

\section{Experiments}
\subsection{Implementation Details}
\label{sec:implementation}
\textbf{Data Preparation.} The evaluated VLIT datasets are clarified in Table~\ref{tab:VLIT_data_statistic}. Specifically, to ensure that each dataset is independent of each other and has no overlapping samples, in DetGPT~\citep{pi2023detgpt}, we remove samples generated from MiniGPT-4~\citep{zhu2023minigpt}; in LLaVAR~\citep{zhang2023llavar}, we remove samples generated from LLaVA~\citep{liu2023visual}. Concentrating on the vision-language field, in LAMM~\citep{yin2023lamm} and Macaw~\citep{lyu2023macaw}, we only use the released image-text data. In addition, the data in LLaVA~\citep{liu2023visual} is divided into three independent ones: LLaVA-Conversation (LLaVACo), LLaVA-Detailed description (LLaVADe) and LLaVA-Reasoning (LLaVARe) for their clear difference. 
\begin{table}[tb!]
\small
	\centering
	\caption{The hyperparameters for instruction tuning using the architecture of InstructBLIP~\citep{dai2023instructblip}, which adopts the Q-Former as the projection module.}
  \label{tab:implement_IBP}
  \begin{tabular}{ p{4.0cm}<{\raggedright}  p{2.0cm}<{\centering} }
    \toprule
    \multicolumn{2}{l}{\textbf{Hyperparameters}}  \\
    \midrule
    Epochs & 5\\
    Warmup Epochs & 1 \\
    Warmup initial learning rate \quad  \quad & 1e-8 \\
    Warmup end learning rate & 1e-5 \\
    Warmup Schedule & Linear \\
    Learning rate decay & Cosine \\
    End (Minimum) learning rate & 0 \\
    Batch size & 128 \\
    Optimizer & AdamW \\
    AdamW $\beta$ & (0.9, 0.999) \\
    Weight decay & 0.05 \\
    \bottomrule
  \end{tabular}
\end{table}
\begin{table}[tb!]
\small
	\centering
	\caption{The hyperparameters for instruction tuning using the architecture of LLaVA~\citep{liu2023visual}, which adopts the linear layer as the projection module.}
  \label{tab:implement_llava}
  \begin{tabular}{ p{4.0cm}<{\raggedright}  p{2.0cm}<{\centering}   }
    \toprule
    \multicolumn{2}{l}{\textbf{Hyperparameters}}  \\
    \midrule
    Epochs & 3\\
    Learning rate & 2e-5 \\
    Learning rate decay & Cosine \\
    Batch size & 128 \\
    Optimizer & AdamW \\
    Weight decay & 0.0 \\
    \bottomrule
  \end{tabular}
\end{table}
\begin{figure*}[tb!]
\centering
\includegraphics[width=14cm]{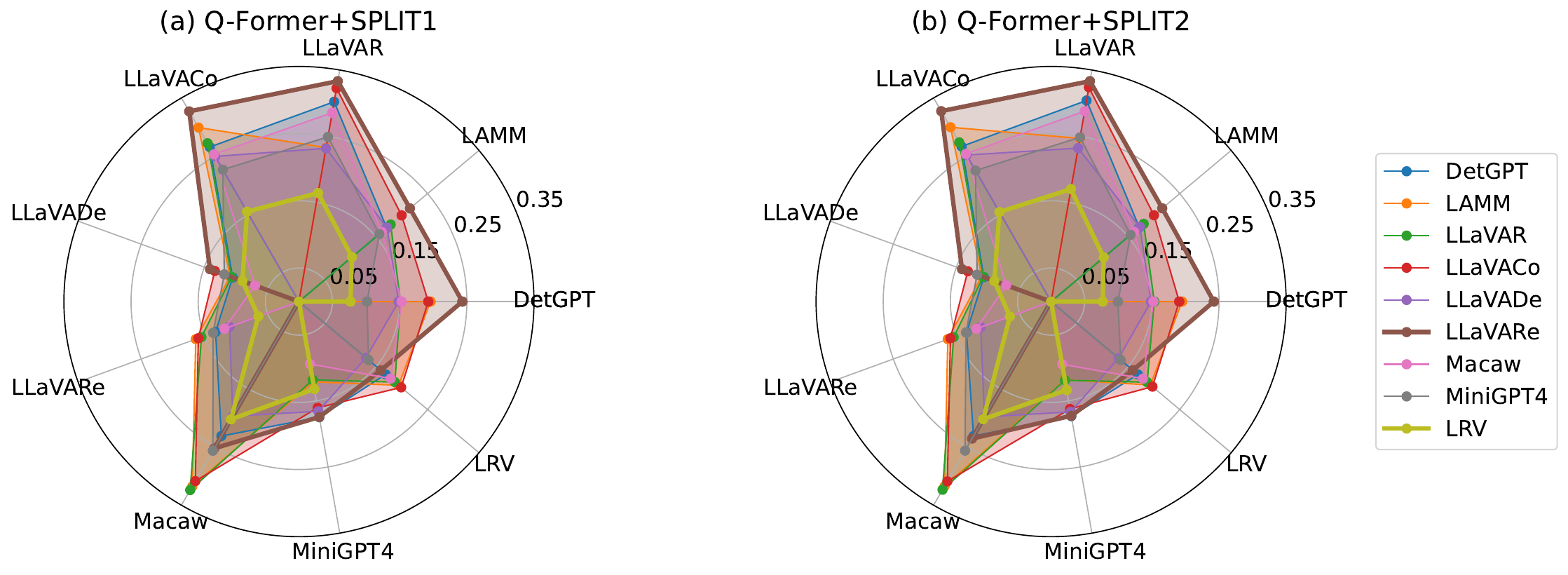}
\caption{Visualizations of $MQ^D_{T \rightarrow i}(i \neq T)$ in dataset quality evaluation. Lines with different colors represent different datasets $D_T$ used for instruction tuning.} 
\label{fig:DQ_Vis}
\end{figure*}
\begin{table*}[tb!]
\footnotesize
	\centering
	\caption{DQ evaluated on SPLIT1 and SPLIT2 by using the Q-Former based architecture.}
  \label{tab:DQ}
  \begin{tabular}{ p{2.2cm}<{\centering} | p{0.8cm}<{\centering} p{0.8cm}<{\centering} p{0.8cm}<{\centering} p{0.8cm}<{\centering} p{0.8cm}<{\centering} p{0.8cm}<{\centering} p{0.8cm}<{\centering} p{1.2cm}<{\centering} p{0.4cm}<{\centering}}
    \toprule
    $D_T$ & DetGPT & LAMM & LLaVAR & LLaVACo & LLaVADe & LLaVARe & Macaw & MiniGPT-4 & LRV  \\
    \midrule
    Q-Former+SPLIT1 & 2.55 & 2.63 & 2.49 & 2.68 & 2.40 & \textbf{\textcolor{red}{2.85}} & 2.31 & 2.38 & \textbf{\textcolor{blue}{1.99}} \\
    Q-Former+SPLIT2 & 2.56 & 2.64 & 2.50 & 2.67 & 2.41 & \textbf{\textcolor{red}{2.83}} & 2.32 & 2.37 & \textbf{\textcolor{blue}{1.99}} \\
    \bottomrule
  \end{tabular}
\end{table*}

To validate the effectiveness of the proposed data refinement strategy, we need to design an evaluation set covering all capabilities of these datasets. For this, we collect $80\%$ samples from each dataset to build independent training sets, on which the \textit{tune-cross-evaluation} paradigm and refinement are performed, and collect 600 samples from the remaining $20\%$ samples to build a balanced and comprehensive evaluation set, namely \emph{Eval600}, as shown in Fig.~\ref{fig:Effect_Val}. We choose 600 samples from each dataset for testing because the smallest dataset MiniGPT-4~\citep{zhu2023minigpt} includes about 3,500 samples, and the $20\%$ includes no more than 700 samples. To build a balanced and comprehensive evaluation set, we finally set the number of selected samples from each dataset for evaluation as 600. To verify the effectiveness of the proposed evaluation paradigm and data refinement on different data partitions, and ensure the universality of experimental effect verification, we perform such data split twice and get two sets, denoted as SPLIT1 and SPLIT2.

\textbf{Instruction Tuning.} In our main experiments, we adopt the architecture of InstructBLIP~\citep{dai2023instructblip} for data evaluation and refinement. The learnable projection module is the Q-Former in BLIP2~\citep{li2023blip}, the vision encoder is the pre-trained ViT-G/14 from EVA-CLIP~\citep{fang2023eva}, and the language model is Vicuna-7B~\citep{vicuna2023}. Specifically, based on the selected vision encoder and language model, the Q-Former used for instruction tuning has been pre-trained on 129M images~\citep{li2023blip}, including COCO~\citep{lin2014microsoft}, Visual Genome~\citep{krishna2017visual}, CC3M~\citep{sharma2018conceptual}, CC12M~\citep{changpinyo2021conceptual}, SBU~\citep{ordonez2011im2text} and LAION400M~\citep{schuhmann2021laion}. Based on the official code of InstructBLIP~\citep{dai2023instructblip}, the learning hyperparameters during instruction tuning are listed in Table~\ref{tab:implement_IBP}. Each dataset has been adopted for tuning on 8 Nvidia A100 (80G) GPUs with the vision encoder and language model kept frozen, only parameters in the Q-Former are optimized. 

In addition, we perform data evaluation and refinement using the architecture of LLaVA~\citep{liu2023visual}, which adopts the linear layer as the projection module for VL alignment. The vision encoder is the pre-trained ViT-L/14 in CLIP~\citep{radford2021learning}, and the language model is Vicuna-7B~\citep{vicuna2023}. The linear layer used for instruction tuning has been pre-trained on 558K image-text pairs from LAION~\citep{schuhmann2021laion}, CC~\citep{sharma2018conceptual} and SBU~\citep{ordonez2011im2text}. We adopt the official code of LLaVA~\citep{liu2023visual} for instruction tuning with their default learning hyperparameters, which are given in Table~\ref{tab:implement_llava}. Each dataset has been adopted for tuning on 8 Nvidia A100 (80G) GPUs with the vision encoder and language model kept frozen, and only parameters in the linear layer are optimized. 

\subsection{DQ Evaluation}
\begin{figure*}[tb!]
\centering
\includegraphics[width=15cm]{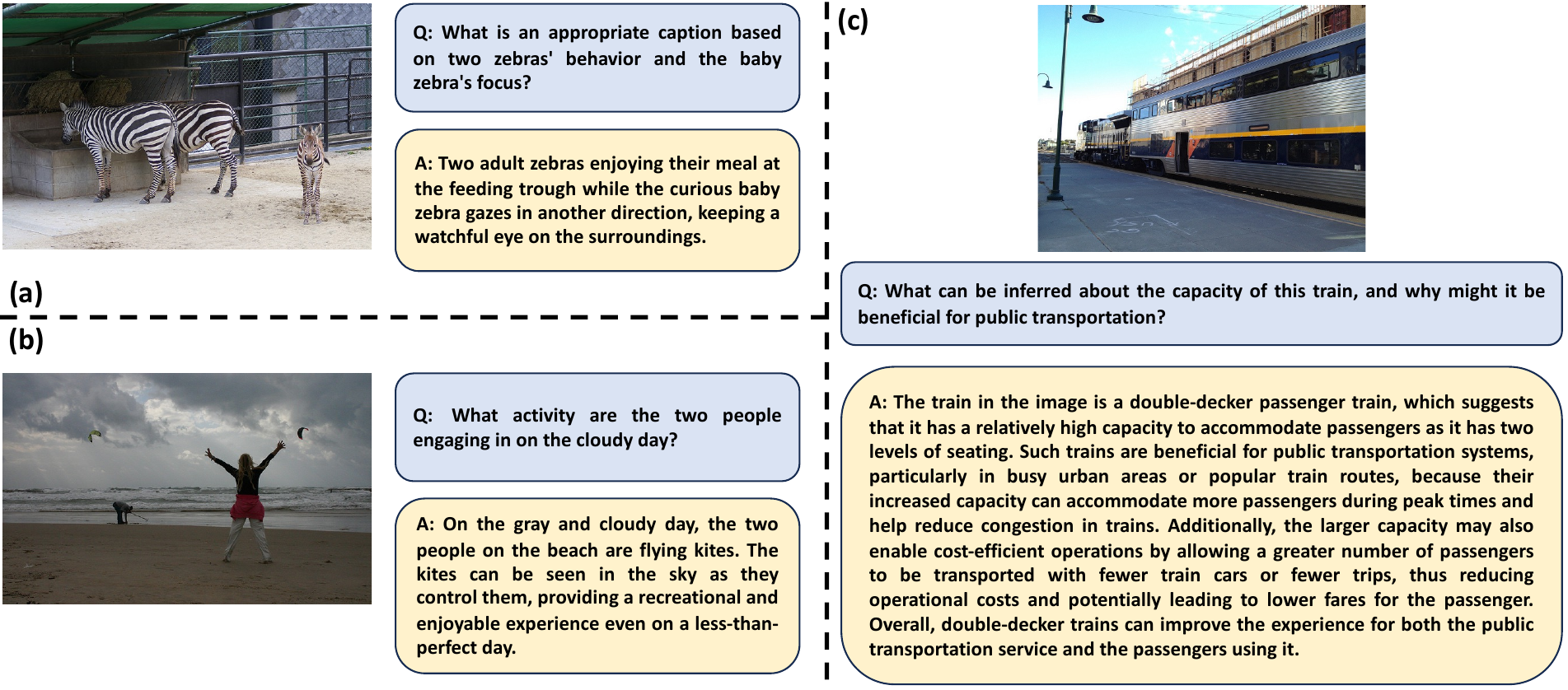}
\caption{Three samples in LLaVA-Reasoning. (a) and (b) are easy reasoning problems, and similar to problems of describing images. (c) is a hard reasoning problem requiring logical thoughts. Q: question. A: answer.} 
\label{fig:LLaVARe}
\end{figure*}

By setting each dataset as the tuning one $D_T$, its one-side qualities measured by other datasets $MQ^D_{T \rightarrow i} (i \neq T)$ are visualized in Fig.~\ref{fig:DQ_Vis}. The areas enclosed by brown and yellow lines are the largest and smallest, indicating that LLaVA-Reasoning and LRV hold the greatest and poorest comprehensive capability. It follows that LLaVA-Reasoning exhibits the highest DQ and LRV shows the lowest DQ among these datasets, as shown in Table~\ref{tab:DQ} computed by Eq.~\ref{eq:DQ}. We infer its reason as that LLaVA-Reasoning includes various problems of which the difficulty varies from low to high. As shown in Fig.~\ref{fig:LLaVARe}, easy reasoning problems may be similar to description problems, while hard reasoning problems may require logical thoughts. As a result, LLaVA-Reasoning exhibits the greatest comprehensive capability. Besides, the results achieved on SPLIT1 and SPLIT2 demonstrate a high degree of consistency, indicating that the DQ evaluation can provide common and objective data analysis. 

\subsection{SQ Evaluation}
\begin{figure*}[tb!]
\centering
\includegraphics[width=12.6cm]{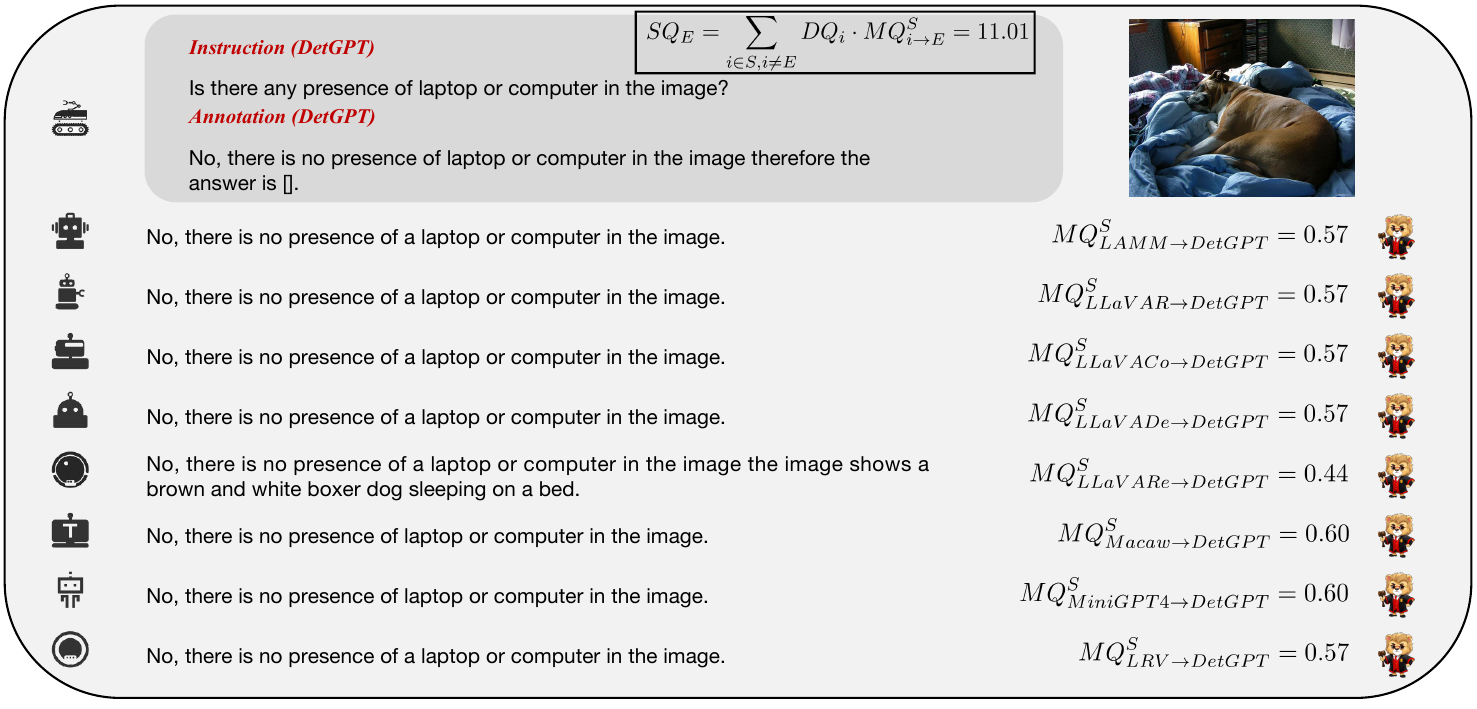}
\caption{A sample in DetGPT with high SQ measured by other datasets.} 
\label{fig:DetGPT_high_main}
\end{figure*}

We show a case of SQ evaluation in Fig.~\ref{fig:DetGPT_high_main}. By calculating $MQ^S_{i \rightarrow E} (i \neq E)$ between the generated answers and the annotation, the comprehensive quality of this sample can be obtained. Due to the high similarity among the generated answers, the calculated SQ of this sample is quite high. More evaluation cases of SQ are delivered in Appendix~\ref{sec:Appen_Evaluation}.

\subsection{Ablation Study on MQ and DQ}
\label{sec:ablate_MQ}
To validate the rationality of the definition of MQ, based on which DQ is devised, we perform ablation studies on 3 combinations of MQ, in which C1 refers Eq.~\ref{eq:MQ}, C2 and C3 are defined as:
\begin{equation}
\begin{aligned}
&C2: MQ = mean (B@4+M+R); \\
&C3: MQ = mean (M+R). 
\label{eq:MQ_ablate}
\end{aligned}
\end{equation}

Given a dataset, its DQ quantified by a reasonable evaluation criteria should be consistent with its performance in the comprehensive evaluation. By setting \emph{Eval600} as the comprehensive evaluation set and $MQ^D_{T\rightarrow Eval600}$ as the performance quantification, according to the three definitions, the results and relative orders of DQ and $MQ^D_{T\rightarrow Eval600}$ achieved on SPLIT1 are shown in Table~\ref{tab:ablate_MQ_DQ_Order}. Compared with C3, C1 and C2 make more consistent results between DQ evaluation and $MQ^D_{T\rightarrow Eval600}$. Because a dataset owning higher DQ should exhibit better all-sided ability, and perform better in the comprehensive evaluation, C1 and C2 are more rational than C3. To preserve a more general evaluation covering as many metrics as possible, we choose C1 as the final definition of MQ, based on which DQ and SQ are devised.
\begin{table*}[tb!]
    \footnotesize
	\centering
	\caption{Ablation study on the definition of MQ. The blue numbers after the results represent their relative rankings. The bold blue numbers indicate the inconsistent ranking relations.}
  \label{tab:ablate_MQ_DQ_Order}
  \begin{tabular}{ p{0.5cm}<{\centering} | p{1.8cm}<{\centering} | p{0.8cm}<{\centering} p{0.8cm}<{\centering} p{0.8cm}<{\centering} p{0.8cm}<{\centering} p{0.8cm}<{\centering} p{0.8cm}<{\centering} p{0.8cm}<{\centering} p{1.2cm}<{\centering} p{0.8cm}<{\centering}}
    \toprule
    \multicolumn{2}{c|}{$D_T$ (Q-Former+SPLIT1)} & DetGPT & LAMM & LLaVAR & LLaVACo & LLaVADe & LLaVARe & Macaw & MiniGPT-4 & LRV  \\
    \midrule
    \multirow{2}{*}{C1} & DQ & 2.55 \textbf{\textcolor{blue}{(4)}} & 2.63 \textbf{\textcolor{blue}{(3)}} & 2.49 \textcolor{blue}{(5)} & 2.68 \textcolor{blue}{(2)} & 2.40 \textcolor{blue}{(6)} & 2.85 \textcolor{blue}{(1)} & 2.31 \textcolor{blue}{(8)} & 2.38 \textcolor{blue}{(7)} & 1.99 \textcolor{blue}{(9)} \\
    & $MQ^D_{T \rightarrow Eval600}$ & 1.37 \textbf{\textcolor{blue}{(3)}} & 1.35 \textbf{\textcolor{blue}{(4)}} & 1.27 \textcolor{blue}{(5)} & 1.42 \textcolor{blue}{(2)} & 1.16 \textcolor{blue}{(6)} & 1.54 \textcolor{blue}{(1)} & 1.11 \textcolor{blue}{(8)} & 1.13 \textcolor{blue}{(7)} & 0.78 \textcolor{blue}{(9)} \\
    \midrule
    \multirow{2}{*}{C2} & DQ & 2.48 \textcolor{blue}{(5)} & 2.61 \textcolor{blue}{(3)} & 2.57 \textcolor{blue}{(4)} & 2.66 \textbf{\textcolor{blue}{(2)}} & 2.38 \textcolor{blue}{(8)} & 2.73 \textbf{\textcolor{blue}{(1)}} & 2.46 \textcolor{blue}{(6)} & 2.40 \textcolor{blue}{(7)} & 2.20 \textcolor{blue}{(9)} \\
    & $MQ^D_{T \rightarrow Eval600}$ & 0.64 \textcolor{blue}{(5)} & 0.70 \textcolor{blue}{(3)} & 0.69 \textcolor{blue}{(4)} & 0.73 \textbf{\textcolor{blue}{(1)}} & 0.57 \textcolor{blue}{(8)} & 0.71 \textbf{\textcolor{blue}{(2)}} & 0.63 \textcolor{blue}{(6)} & 0.59 \textcolor{blue}{(7)} & 0.49 \textcolor{blue}{(9)} \\
    \midrule
    \multirow{2}{*}{C3} & DQ & 2.79 \textcolor{blue}{(6)} & 2.95 \textcolor{blue}{(3)} & 2.94 \textbf{\textcolor{blue}{(4)}} & 3.00 \textbf{\textcolor{blue}{(2)}} & 2.72 \textcolor{blue}{(8)} & 3.05 \textbf{\textcolor{blue}{(1)}} & 2.83 \textcolor{blue}{(5)} & 2.72 \textcolor{blue}{(7)} & 2.59 \textcolor{blue}{(9)} \\
    & $MQ^D_{T \rightarrow Eval600}$ & 0.50 \textcolor{blue}{(6)} & 0.56 \textcolor{blue}{(3)} & 0.57 \textbf{\textcolor{blue}{(2)}} & 0.59 \textbf{\textcolor{blue}{(1)}} & 0.47 \textcolor{blue}{(8)} & 0.54 \textbf{\textcolor{blue}{(4)}} & 0.53 \textcolor{blue}{(5)} & 0.48 \textcolor{blue}{(7)} & 0.44 \textcolor{blue}{(9)} \\
    \bottomrule
  \end{tabular}
\end{table*}
\begin{table*}[tb!]
    \footnotesize
	\centering
	\caption{$MQ^D_{T \rightarrow Eval600}$ on SPLIT1 and SPLIT2 by using the Q-Former based architecture.}
  \label{tab:single_vs_merge}
  \begin{tabular}{ p{2.2cm}<{\centering} | p{0.7cm}<{\centering} p{0.7cm}<{\centering} p{0.8cm}<{\centering} p{0.8cm}<{\centering} p{0.8cm}<{\centering} p{0.8cm}<{\centering} p{0.7cm}<{\centering} p{1.2cm}<{\centering} p{0.5cm}<{\centering}  p{0.7cm}<{\centering}}
    \toprule
    $D_T$ & DetGPT & LAMM & LLaVAR & LLaVACo & LLaVADe & LLaVARe & Macaw & MiniGPT-4 & LRV & Merge  \\
    \midrule
    Q-Former+SPLIT1 & 1.37 & 1.35 & 1.27 & 1.42 & 1.16 & 1.54 & 1.11 & 1.13 & 0.78 & \textbf{1.64} \\
    Q-Former+SPLIT2 & 1.38 & 1.36 & 1.29 & 1.43 & 1.18 & 1.55 & 1.12 & 1.12 & 0.79 & \textbf{1.64} \\
    \bottomrule
  \end{tabular}
\end{table*}

\subsection{Single Dataset VS. Merged Dataset}
\label{sec:single_vs_merge}
To build a comprehensive dataset integrating all capabilities, a simple yet direct way is to add all these single datasets together into one, denoted as ``Merge". By setting \emph{Eval600} as the evaluation dataset, the $MQ^D_{T\rightarrow Eval600}$ achieved by setting each single dataset and the merged dataset as tuning one $D_T$ is compared in Table~\ref{tab:single_vs_merge}. The simply merged dataset achieves the greatest result, showing adding all datasets together can contribute to an all-powerful model that exhibits the best performance on the comprehensive evaluation set covering all capabilities. 

\subsection{REVO-LION and Ablation Study on Refinement Strategy}
\label{sec:ablate_refine}
It has been validated that combining all datasets together can develop an all-powerful model in a comprehensive evaluation compared with single datasets in Sec.~\ref{sec:single_vs_merge}. Considering that data quality is more significant than data quantity~\citep{zeng2023matters}, we further perform data refinement based on the above holistic evaluation. Specifically, we collect part of the samples from each dataset to build a comprehensive dataset. In addition to the refinement strategy defined in Eq.~\ref{eq:revolion}, denoted as $S1$, we design another two strategies for comparisons. The second strategy, namely $S2$, collects the samples from each dataset randomly with the same amount as in $S1$. The third strategy $S3$ adopts the Gaussian distribution for sample selection. Specifically, for each dataset $D_i (i \in S)$, we calculate the mean value $\mu_i$ and the standard deviation $\sigma_i$ of SQ of the samples in $D_i$. The sample whose SQ exists within an interval of $\lambda$ times the standard deviation $\sigma_i$ around the mean value $\mu_i$ will be selected. $S3$ is formulated as:
\begin{equation}
\begin{aligned}
S3 &= \bigcup\limits_{i \in S} \boldsymbol{x}^k_i, \\ (\boldsymbol{x}^k_i \in D_i, SQ^k_i \in [&\mu_i - \lambda \cdot \sigma_i, \mu_i + \lambda \cdot \sigma_i]).
\label{eq:revolion_S3}
\end{aligned}
\end{equation}

We adopt CIDEr, the hold-out metric in defining MQ, to measure the comprehensive performance of the model tuned on the refined dataset, thus making an objective evaluation for the data refinement. By setting \emph{Eval600} as the comprehensive evaluation set, and selecting a portion $P$ of samples in each dataset, the result comparisons between $S1$ and $S2$ are given in Table~\ref{tab:S1_S2_Refine}. The case when $P=100\%$ refers to simply adding all datasets together. 
\begin{table*}[tb!]
    \footnotesize
	\centering
	\caption{Evaluation on \emph{Eval600} measured by CIDEr using refinement strategies $S1$ and $S2$ with the portion $P$ ranging from $10\%$ to $100\%$. ``Nums" refers to the number of image-instruction-answer triplets, that are selected by our scheme from raw datasets, in the refined dataset for tuning. Compared with the results marked with gray achieved by all data, the results marked with yellow achieved with half of the full data are competitive, and the results marked with red are all better.}
  \label{tab:S1_S2_Refine}
  \begin{tabular}{ p{2.1cm}<{\centering} | p{1.6cm}<{\centering} | p{0.6cm}<{\centering} p{0.6cm}<{\centering} p{0.6cm}<{\centering} p{0.6cm}<{\centering} p{0.6cm}<{\centering} p{0.6cm}<{\centering} p{0.6cm}<{\centering} p{0.6cm}<{\centering} p{0.6cm}<{\centering} | p{0.6cm}<{\centering}}
    \toprule
    \multicolumn{2}{c|}{Portion ($P$)} & $10\%$ & $20\%$ & $30\%$ & $40\%$ & $50\%$ & $60\%$ & $70\%$ & $80\%$ & $90\%$ &  $100\%$  \\
    \midrule
    \multirow{3}{*}{Q-Former+SPLIT1} & Nums & 92828 & 185650 & 278473 & 371293 & 464115 & 556940 & 649760 & 742584 & 835406 & 928225\\
    & $S1$-CIDEr & 163.43 & 168.56 & 171.64 & 174.54 & \cellcolor{yellow}175.13 & \cellcolor{pink}177.21 & \cellcolor{pink}194.70 & \cellcolor{pink}177.16 & \cellcolor{pink}176.64 & \cellcolor{lightgray}175.49 \\
    & $S2$-CIDEr & 165.21 & 168.63 & 170.18 & 171.91 & 172.82 & 174.37 & 174.22 & 175.62 & 176.03 & 175.49 \\
    \midrule
    \multirow{3}{*}{Q-Former+SPLIT2} & Nums & 92807 & 185608 & 278410 & 371211 & 464012 & 556815 & 649616 & 742418 & 835219 & 928017 \\
    & $S1$-CIDEr & 165.03 & 170.87 & 173.56 & 174.89 & \cellcolor{yellow}175.99 & \cellcolor{pink}178.33 & \cellcolor{pink}178.87 & \cellcolor{pink}178.23 & \cellcolor{pink}178.80 & \cellcolor{lightgray}175.49 \\
    & $S2$-CIDEr & 165.81 & 169.49 & 172.40 & 173.89 & 175.78 & 176.23 & 177.16 & 178.23 & 179.17 & 175.49 \\
    \bottomrule
  \end{tabular}
\end{table*}

For the refinement strategy S1, the results when $P \in [50\%, 90\%]$ are all competitive and even better than those when using the simply merged dataset. It shows that S1 successfully collects high-quality samples in the refined dataset. Specifically, the CIDEr rises with the increase of $P$ from $10\%$ to $70\%$. When we select the top $50\%$ samples with higher SQ from each dataset, we can already achieve competitive performance comparable to those using the entire data. Then, the CIDEr achieves the highest when $P=70\%$. When $P$ increases from $70\%$ to $100\%$, the CIDEr results decrease, which is caused by the involvement of samples with lower SQ. Besides, comparing S1 with $S2$, when keeping the number of collected samples from each dataset the same, results achieved by selecting samples with higher SQ are almost better than those achieved by random selection, which validates that S1 is more effective than $S2$. Moreover, for the refinement strategy $S2$, the CIDEr rises with the increase of $P$ from $10\%$ to $90\%$. It demonstrates that with the lack of effective data evaluation and refinement strategies, a direct way for improving the performance is just expanding the scale of datasets.
\begin{table}[tb!]
        \footnotesize
	\centering
	\caption{Evaluation on \emph{Eval600} measured by CIDEr using the refinement strategy $S3$ by setting $\lambda \in [1.0, 1.5, 2.0]$. ``Nums" refers to the number of image-instruction-answer triplets.}
  \label{tab:Gauss_Refine}
  \begin{tabular}{p{2.1cm}<{\centering} | p{0.8cm}<{\centering} | p{0.6cm}<{\centering} p{0.6cm}<{\centering} p{0.6cm}<{\centering}}
    \toprule
    \multicolumn{2}{c|}{Times ($\lambda$)} & 1.0 & 1.5 & 2.0 \\
    \midrule
    \multirow{2}{*}{Q-Former+SPLIT1} & Nums & 697374 & 838771 & 880426 \\
    & CIDEr & 173.94 & 175.07 & 176.52 \\
    \midrule
    \multirow{2}{*}{Q-Former+SPLIT2} & Nums & 697346 & 838650 & 880206 \\
    & CIDEr & 175.88 & 178.45 & 179.32 \\
    \bottomrule
  \end{tabular}
\end{table}

In addition, results from the refinement strategy $S3$ when setting the times $\lambda$ within [1.0, 1.5, 2.0] are given in Table~\ref{tab:Gauss_Refine}. Comparing the results achieved by setting $\lambda=1.0$ in $S3$ with those achieved by setting $P=70\%$ in S1 in Table~\ref{tab:S1_S2_Refine}, though more samples are collected in $S3$, the performance achieved by S1 with fewer samples is better. The same phenomenon also happens in the comparison between setting $\lambda \in [1.5, 2.0]$ in $S3$ and setting $P=90\%$ in S1 in Table~\ref{tab:S1_S2_Refine}. The comparisons prove that S1 is more effective than $S3$. According to the ablation studies, the effectiveness of the proposed \textit{tune-cross-evaluation} paradigm and the refinement strategy is systematically validated.

\subsection{Data Evaluation and Refinement using the Linear Projection Module}
\label{sec:llava_results}
\begin{figure*}[tb!]
\begin{minipage}[b]{7cm}
  \centering
  \centerline{\includegraphics[width=6cm]{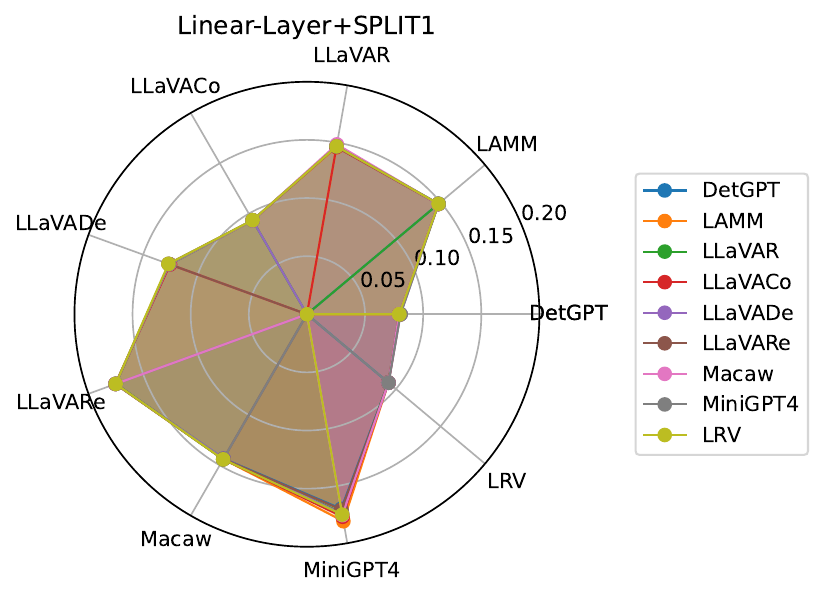}}
\end{minipage}
\hfill
\begin{minipage}[b]{9cm}
  \centering
  \centerline{\includegraphics[width=9cm]{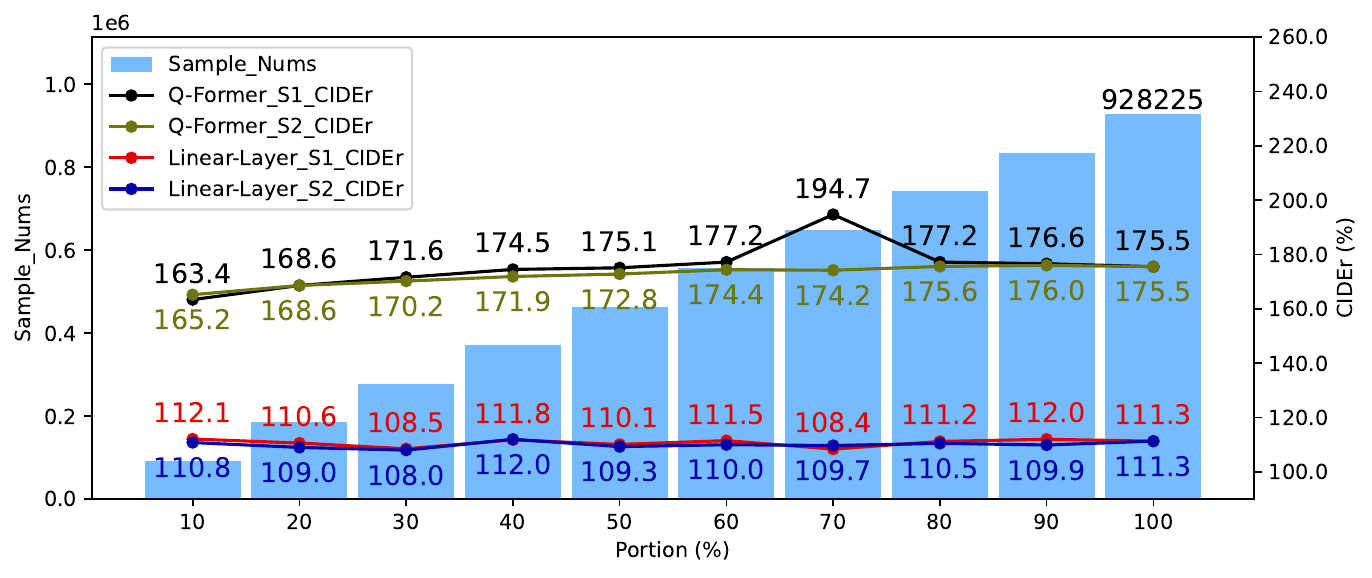}}
\end{minipage}
\caption{Results of data evaluation and refinement using the linear projection-based architecture. (Left) Visualizations of $MQ^D_{T \rightarrow i}(i \neq T)$ in DQ evaluation using the linear projection module. (Right) Result comparisons between using the Q-Former andthe linear layer for projection using strategies $S1$ and $S2$.} 
\label{fig:LLaVA_Res}
\end{figure*}

For a supplementary, we perform the data evaluation and refinement using SPLIT1 based on the architecture adopting the linear layer as the projection module for VL alignment. Specifically, we take the architecture of LLaVA~\citep{liu2023visual}, and its setting has been illustrated in Sec~\ref{sec:implementation}. When setting each dataset as the tuning one $D_T$, its one-side qualities $MQ^D_{T \rightarrow i}(i \neq T)$ measured by other datasets are given in Fig.~\ref{fig:LLaVA_Res} (Left). It shows that each dataset exhibits extremely high similarity in the dataset-wise evaluation, leading to almost equal DQ for each dataset, compared with the results in Fig.~\ref{fig:DQ_Vis}. Consequently, using the linear projection-based VLIT model cannot effectively distinguish differences among datasets, resulting in invalid data evaluation. In addition, based on the evaluation, we perform the data refinement using strategies $S1$ and $S2$. The refinement results achieved by using the Q-Former and the linear layer for projection are shown in Fig.~\ref{fig:LLaVA_Res} (Right). Obviously, when keeping both the vision encoder and the language model frozen, using the linear projection module results in a much more unsatisfying performance than using the Q-Former. Then, taking a deep comparison between the results achieved by S1 and $S2$ using the linear layer-based architecture, the CIDEr results vary within a small range when the portion $P$ of selected samples in each dataset changes. The highest result in S1 refinement, which is higher than using all the data, is achieved when only collecting $10\%$ of samples with higher SQ from each dataset. It shows that as a much simpler projection module, the linear layer does not need as much high-quality instruction data as the Q-Former. The simplicity of the projection module limits the greatest performance that can be improved by expanding the data scale. Besides, compared with $S2$, the strategy $S1$ is almost better with different portions.
\begin{figure*}[tb!]
\centering
\includegraphics[width=15cm]{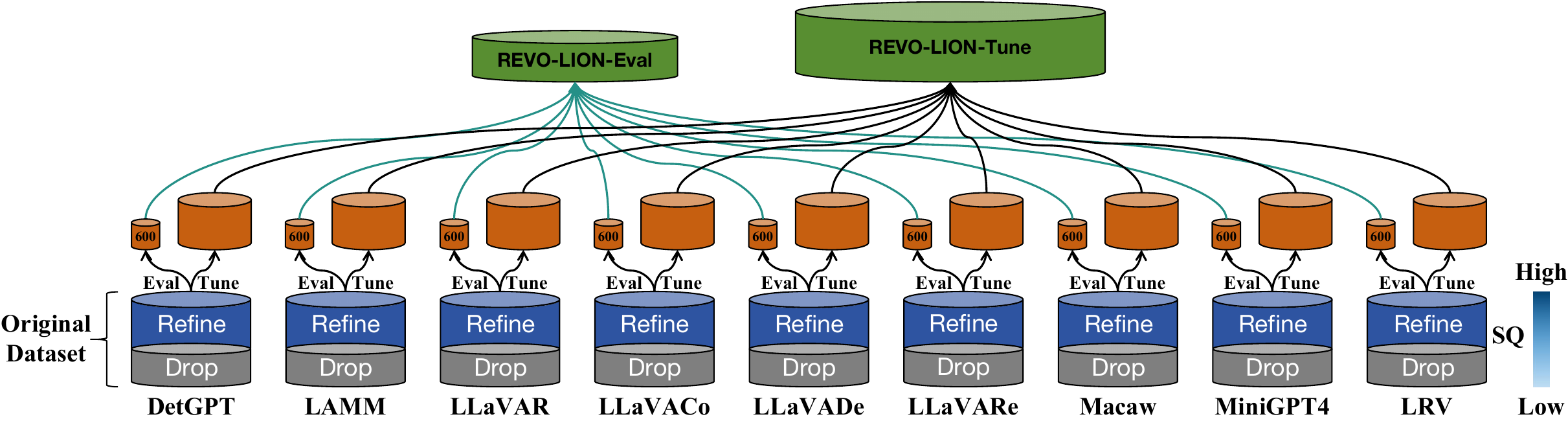}
\caption{The refining process of buiding REVO-LION from existing VLIT datasets. The proposed \textit{tune-cross-evaluation} paradigm is directly performed on each original dataset without partition. After the holistic evaluation, the top $70\%$ samples with higher SQ in each dataset are collected, in which 600 samples are collected into the balanced and comprehensive evaluation benchmark, namely REVO-LION-Eval, and the remaining are collected into the refined tuning dataset, namely REVO-LION-Tune, for developing an all-powerful model.} 
\label{fig:REVO_LION_Fig}
\end{figure*}

Except for the effectiveness of $S1$, which has been validated compared with $S2$, other results are inconsistent with ones using the architecture of InstructBLIP, and the linear projection module is not as good as the Q-Former. We make the deep analysis as follows. (1) From the perspective of the architecture, linear projection is quite simple in transferring the visual feature to the language space. While Q-Former adopts the pre-trained BERT~\citep{kenton2019bert} as initialization, and extracts the desired visual feature according to the texts using a more sophisticated cross-attention mechanism. (2) From the perspective of the pre-trained dataset, both the linear layer and the Q-Former have been pre-trained on large-scale image-text pairs for VL alignment before instruction tuning. As demonstrated in Sec~\ref{sec:implementation}, Q-Former in BLIP2~\citep{li2023blip} has been pre-trained on 129M images~\citep{li2023blip} from COCO~\citep{lin2014microsoft}, Visual Genome~\citep{krishna2017visual}, CC3M~\citep{sharma2018conceptual}, CC12M~\citep{changpinyo2021conceptual}, SBU~\citep{ordonez2011im2text} and LAION400M~\citep{schuhmann2021laion}. While the linear layer in LLaVA~\citep{liu2023visual} has been pre-trained only on 558K image-text pairs from LAION~\citep{schuhmann2021laion}, CC~\citep{sharma2018conceptual} and SBU~\citep{ordonez2011im2text}. The significant difference between the amount of pre-training dataset results in a much poorer VL alignment of the linear projection than the Q-Former.

\section{REVO-LION Release}
According to the validated effectiveness on specific data preparation in above experiments, to release the REVO-LION, the evaluation and refinement are performed on the original datasets without partitions. As analyzed in Sec.~\ref{sec:llava_results}, using the linear layer as the projection module for VL alignment is inferior to using the Q-Former. Therefore, we adopt the architecture of InstructBLIP, the detailed setting of which is delivered in Sec~\ref{sec:implementation}, for data evaluation and refinement.

In the released dataset REVO-LION, the \textit{tune-cross-evaluation} paradigm is directly performed on each original dataset without partition, as shown in Fig~\ref{fig:REVO_LION_Fig}. According to the results in Table~\ref{tab:S1_S2_Refine}, setting the portion $P=70\%$ can achieve the best performance. Therefore, we release the dataset with setting $P=70\%$. After refining each dataset, we divide it into a train set for instruction tuning and an evaluation set as a convenient yet stable benchmark. To keep a balanced dataset for evaluation, we select 600 samples from each refined dataset to build the evaluation set, namely REVO-LION-Eval. The remaining samples in each refined dataset are combined into the instruction tuning dataset, namely REVO-LION-Tune. Moreover, as the annotations in REVO-LION share the same style of open-ended texts generated by LLMs, the caption metrics can be directly adopted for a model-free and human-free performance evaluation when using the REVO-LION-Eval as the benchmark.

\section{Conclusions and Outlook}
In this paper, we pioneer the analysis of VLIT datasets and propose the \textit{tune-cross-evaluation} paradigm. The key idea is to fully use the datasets as both tuning and evaluation sets, by which the ideal annotations consistent with open-ended generated texts are available. Then, we define the Meta Quality (MQ), a model-free and human-free metric, as the mean score measured by caption metrics, including BLEU, METEOR and ROUGE-L. We extend MQ to Dataset Quality (DQ) and Sample Quality (SQ) for quantitative evaluation. Based on the holistic evaluation, we build a refined dataset REVO-LION by collecting samples with higher SQ from each dataset. With only half of the full data, the model trained on the refined dataset can achieve competitive performance compared to that trained on all data. In the released version, REVO-LION includes a train set, which can be commonly used for developing an all-powerful model, and an evaluation set, which can serve as a convenient yet stable benchmark. 

The evaluation paradigm is not limited to the datasets analyzed in this paper. The more datasets with various capabilities are involved in the evaluation, the more comprehensive analysis is achieved. As a result, the refined dataset can be used to develop a VLIT model performing well in more aspects, and also as a more comprehensive evaluation benchmark.

\bmhead{Acknowledgments}
This work was partly supported at the university side, by National Science Foundation of China (62222607, 92370201, 62002252), and Shanghai Municipal Science and Technology Major Project (2021SHZDZX0102).

\bmhead{Data Availability}
There is no new dataset included in this paper. The datasets in our experiments are all publicly available, they are: DetGPT~\citep{pi2023detgpt}, LAMM~\citep{yin2023lamm}, LLaVAR~\citep{zhang2023llavar}, LLaVA~\citep{liu2023visual}, Macaw~\citep{lyu2023macaw}, MiniGPT4~\citep{zhu2023minigpt} and LRV~\citep{liu2023aligning}. 
\bibliography{sn-bibliography}

\appendix
\section{SQ Evaluation Cases}
\label{sec:Appen_Evaluation}
\begin{figure*}[htb]
\centering
\includegraphics[width=12.6cm]{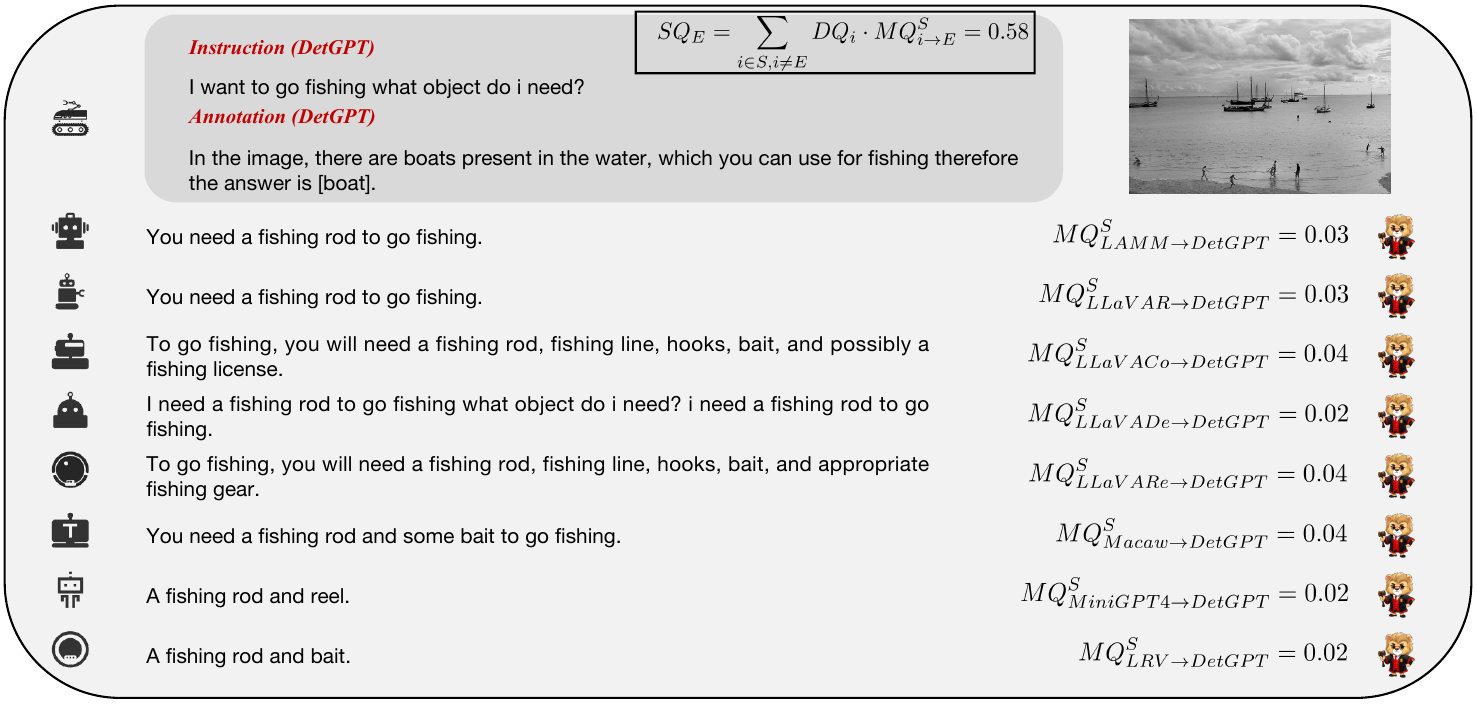}
\caption{A sample in DetGPT with low SQ measured by other datasets.} 
\label{fig:DetGPT_low}
\end{figure*}

\begin{figure*}[htb]
\centering
\includegraphics[width=12.6cm]{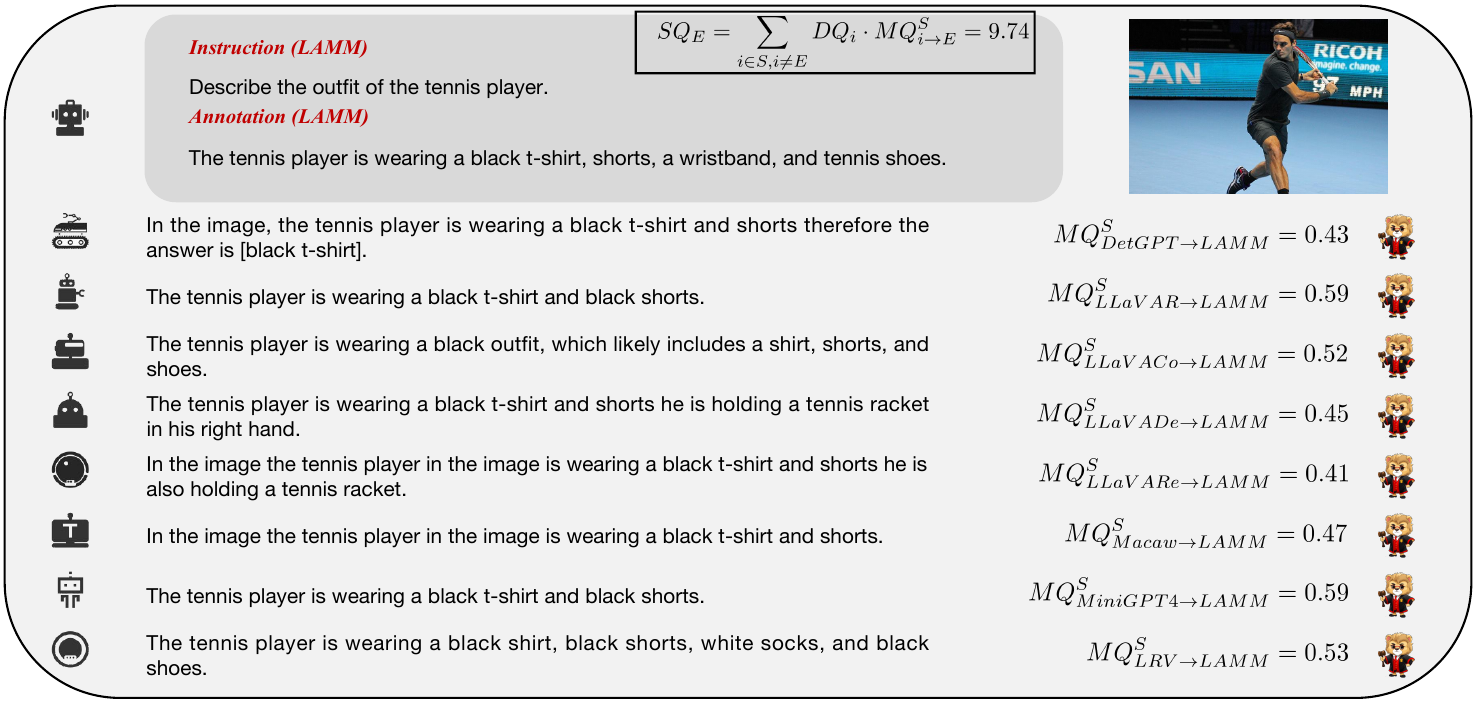}
\caption{A sample in LAMM with high SQ measured by other datasets.} 
\label{fig:LAMM_high}
\end{figure*}

\begin{figure*}[tb!]
\centering
\includegraphics[width=12.6cm]{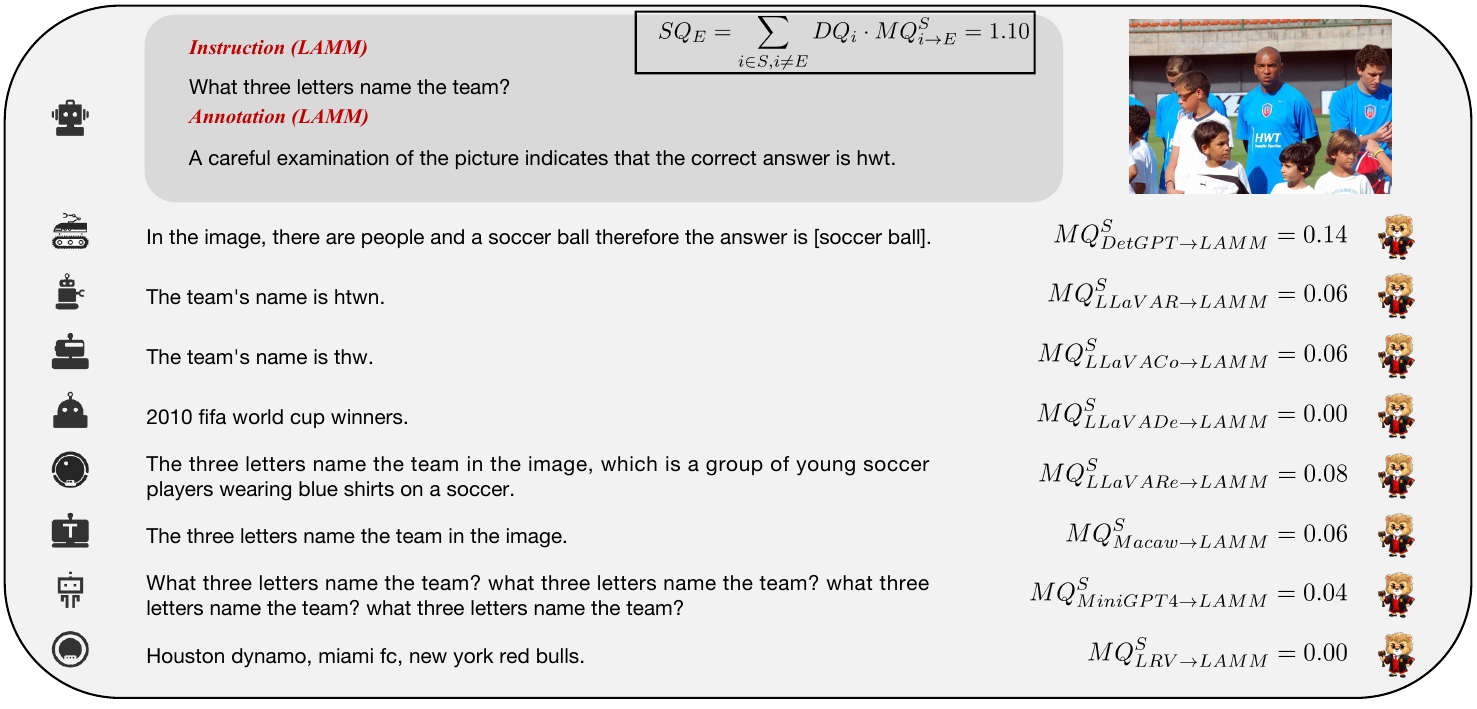}
\caption{A sample in LAMM with low SQ measured by other datasets.} 
\label{fig:LAMM_low}
\end{figure*}

\begin{figure*}[tb!]
\centering
\includegraphics[width=12.6cm]{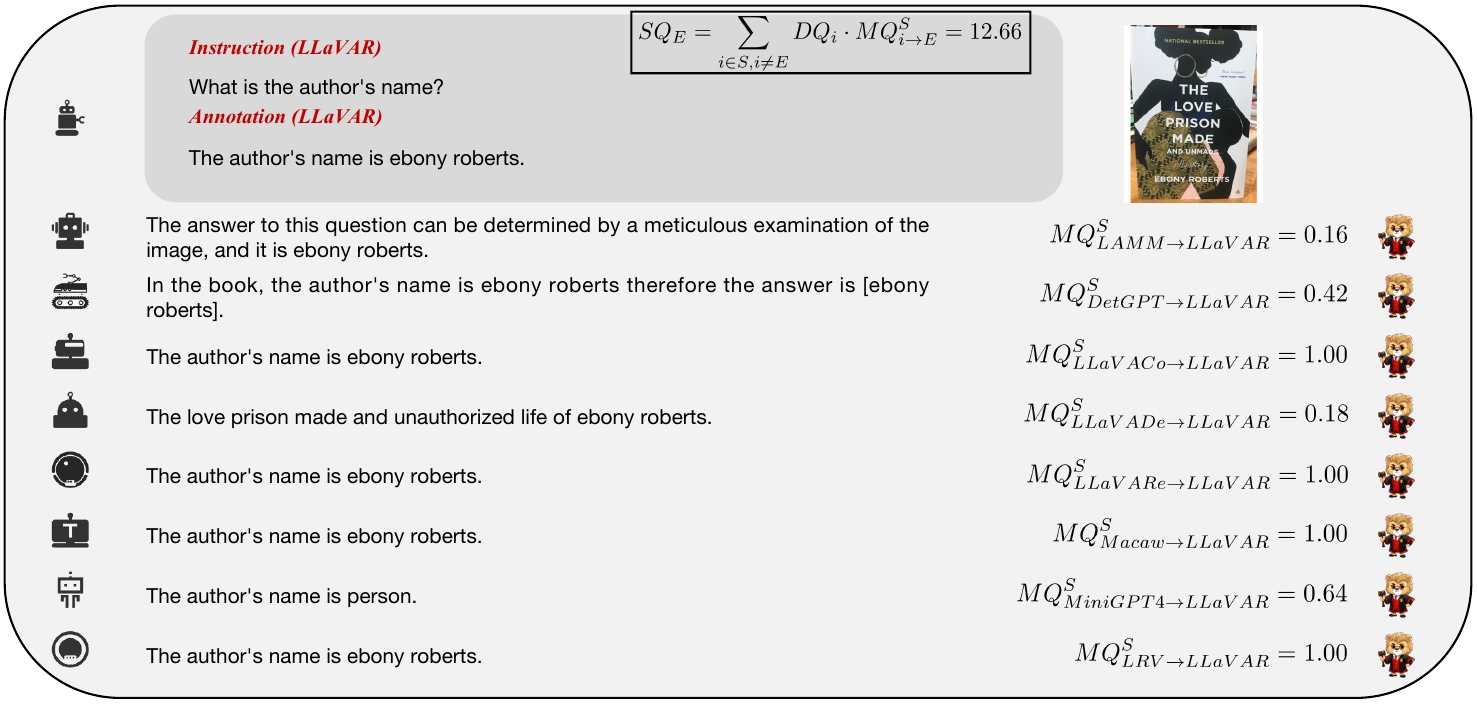}
\caption{A sample in LLaVAR with high SQ measured by other datasets.} 
\label{fig:LLaVAR_high}
\end{figure*}

\begin{figure*}[tb!]
\centering
\includegraphics[width=12.6cm]{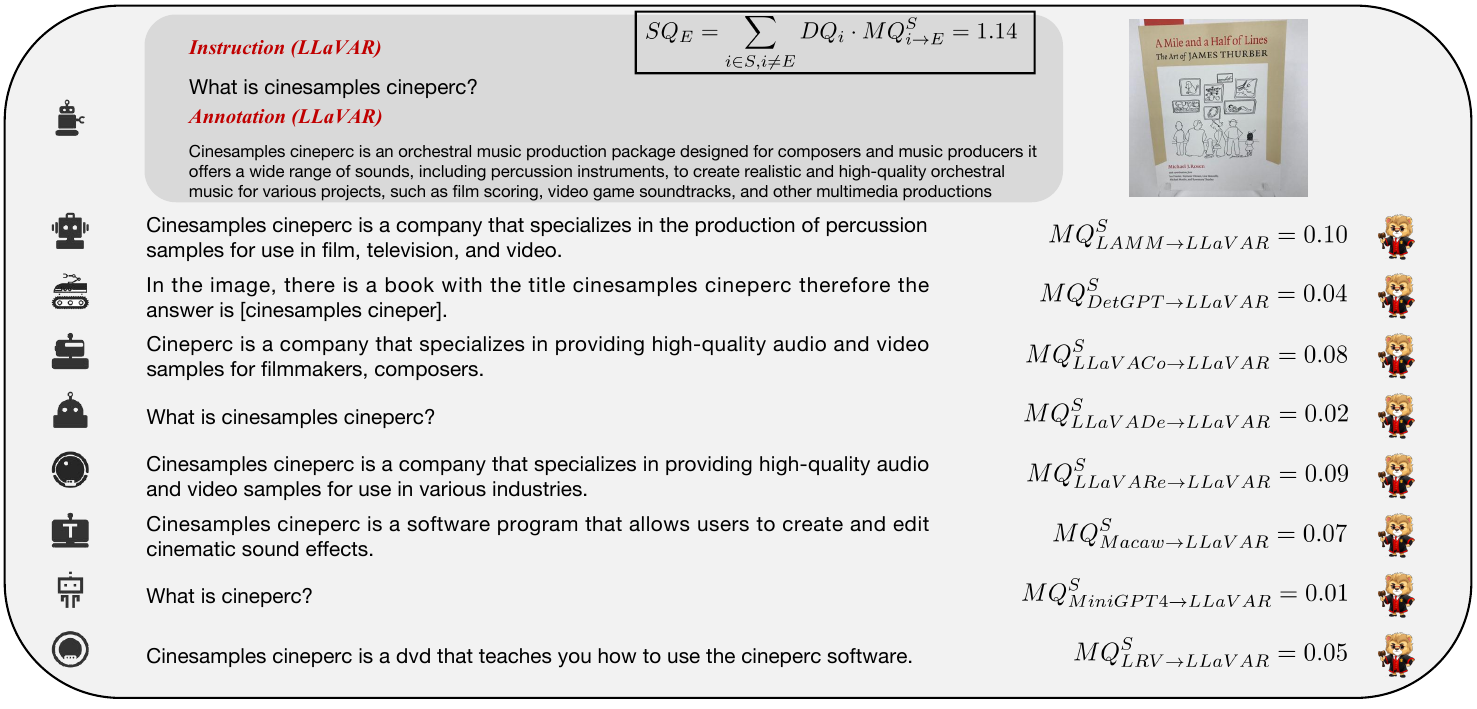}
\caption{A sample in LLaVAR with low SQ measured by other datasets.} 
\label{fig:LLaVAR_low}
\end{figure*}

\begin{figure*}[tb!]
\begin{minipage}[b]{16cm}
  \centering
  \centerline{\includegraphics[width=12.6cm]{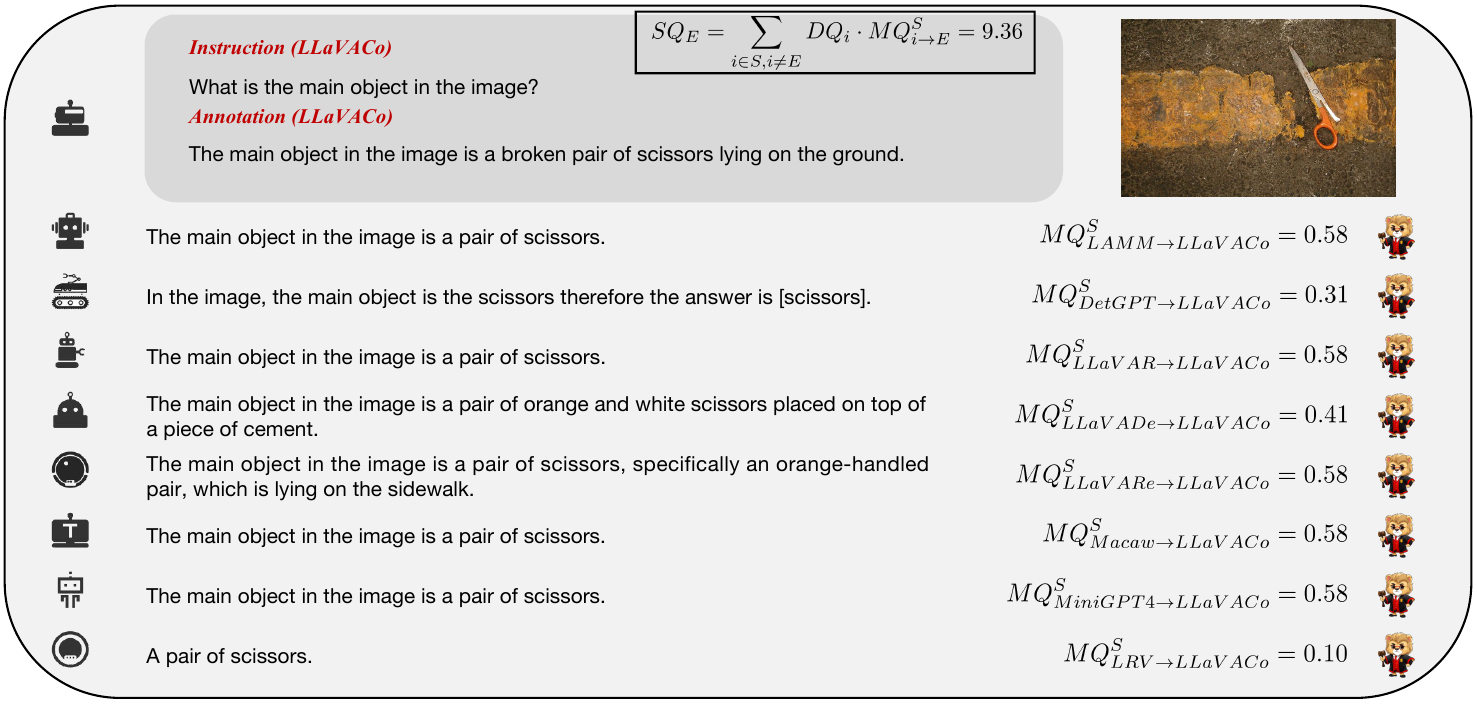}}
  \centerline{}
\end{minipage}

\begin{minipage}[b]{16cm}
  \centering
  \centerline{\includegraphics[width=12.6cm]{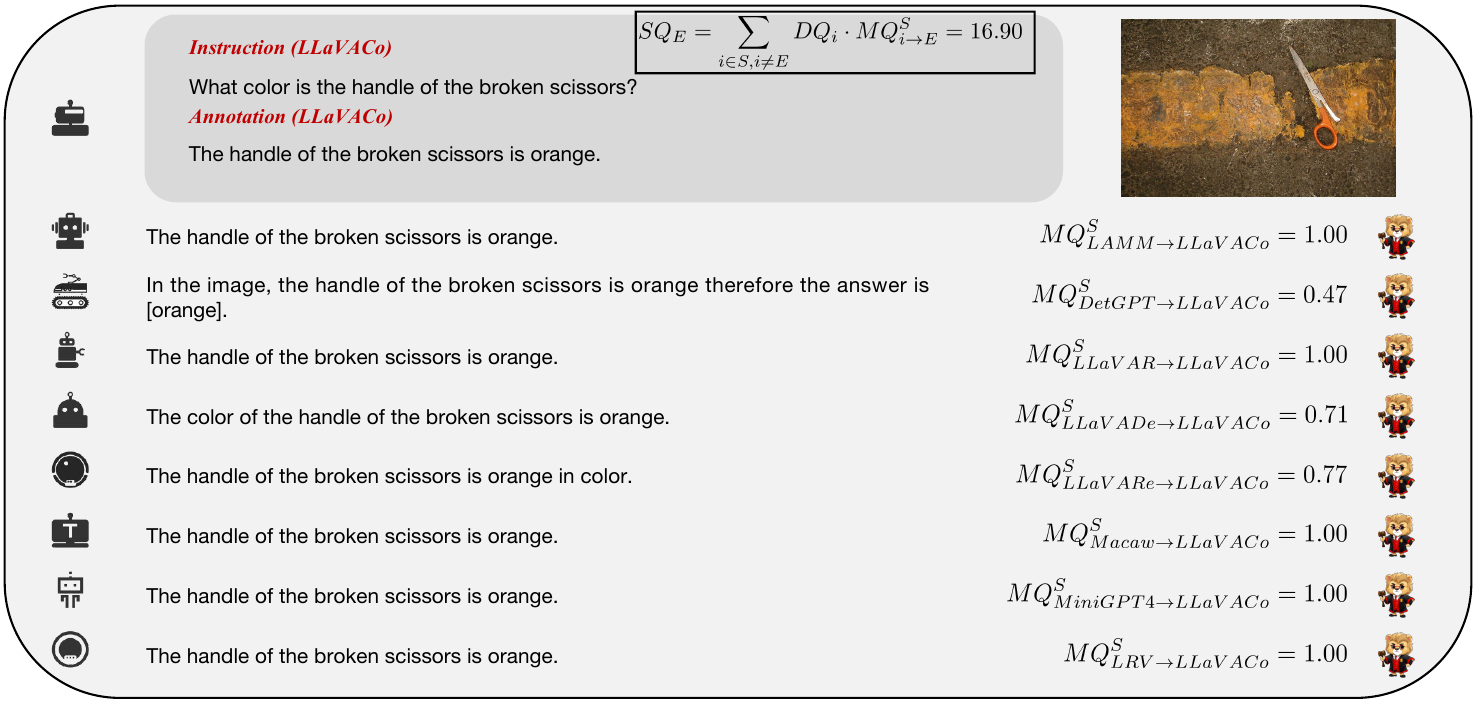}}
  \centerline{}
\end{minipage}

\begin{minipage}[b]{16cm}
  \centering
  \centerline{\includegraphics[width=12.6cm]{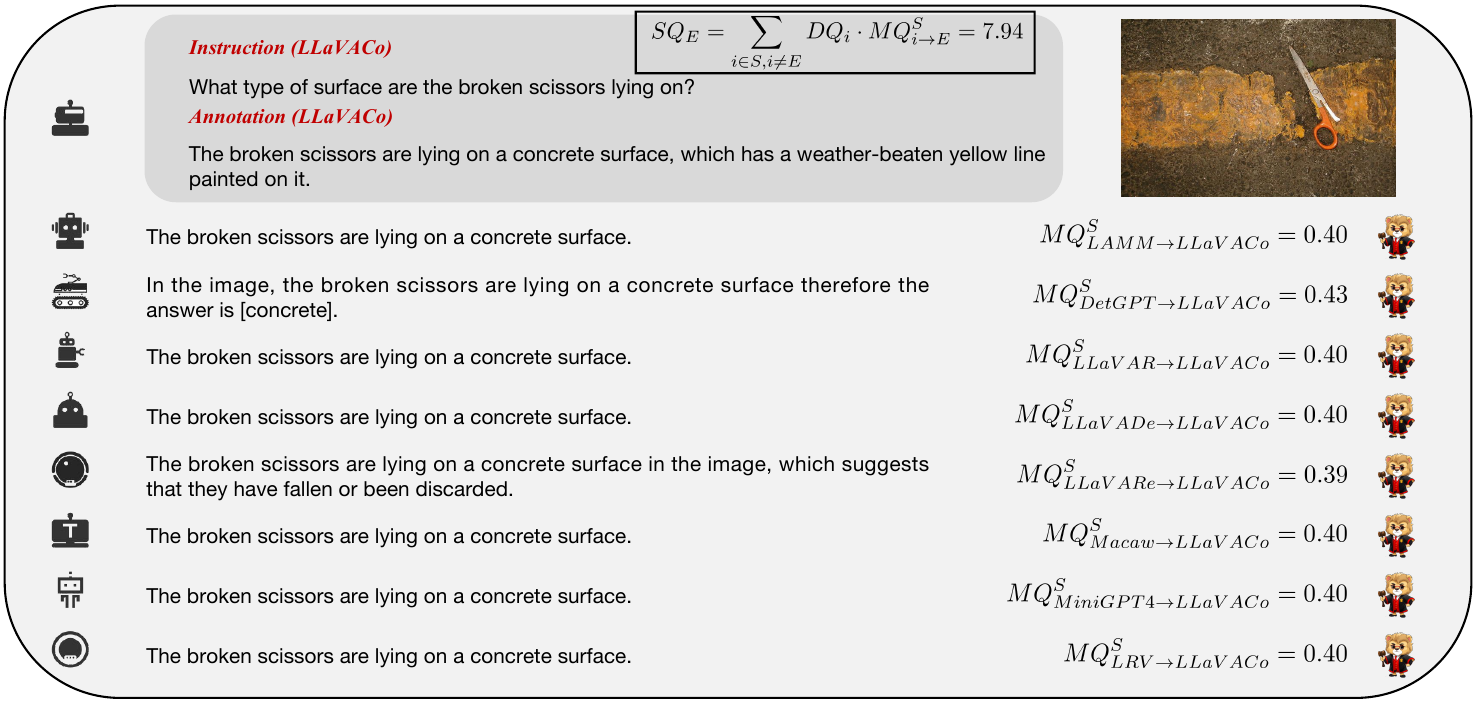}}
  \centerline{}
\end{minipage}
\caption{A conversation sample in LLaVA-Conversation with high SQ measured by other datasets.} 
\label{fig:LLaVACo_high}
\end{figure*}

\begin{figure*}[tb!]
\begin{minipage}[b]{16cm}
  \centering
  \centerline{\includegraphics[width=12.6cm]{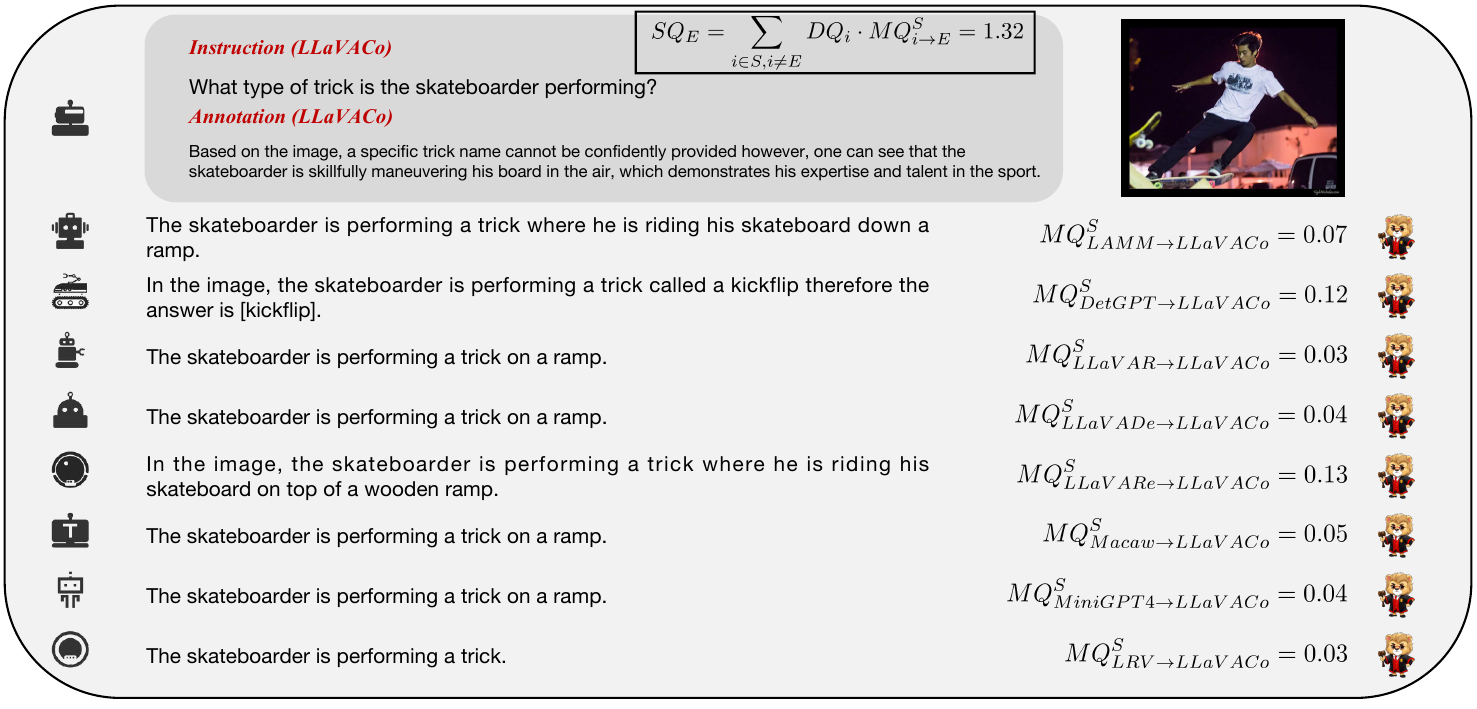}}
  \centerline{}
\end{minipage}

\begin{minipage}[b]{16cm}
  \centering
  \centerline{\includegraphics[width=12.6cm]{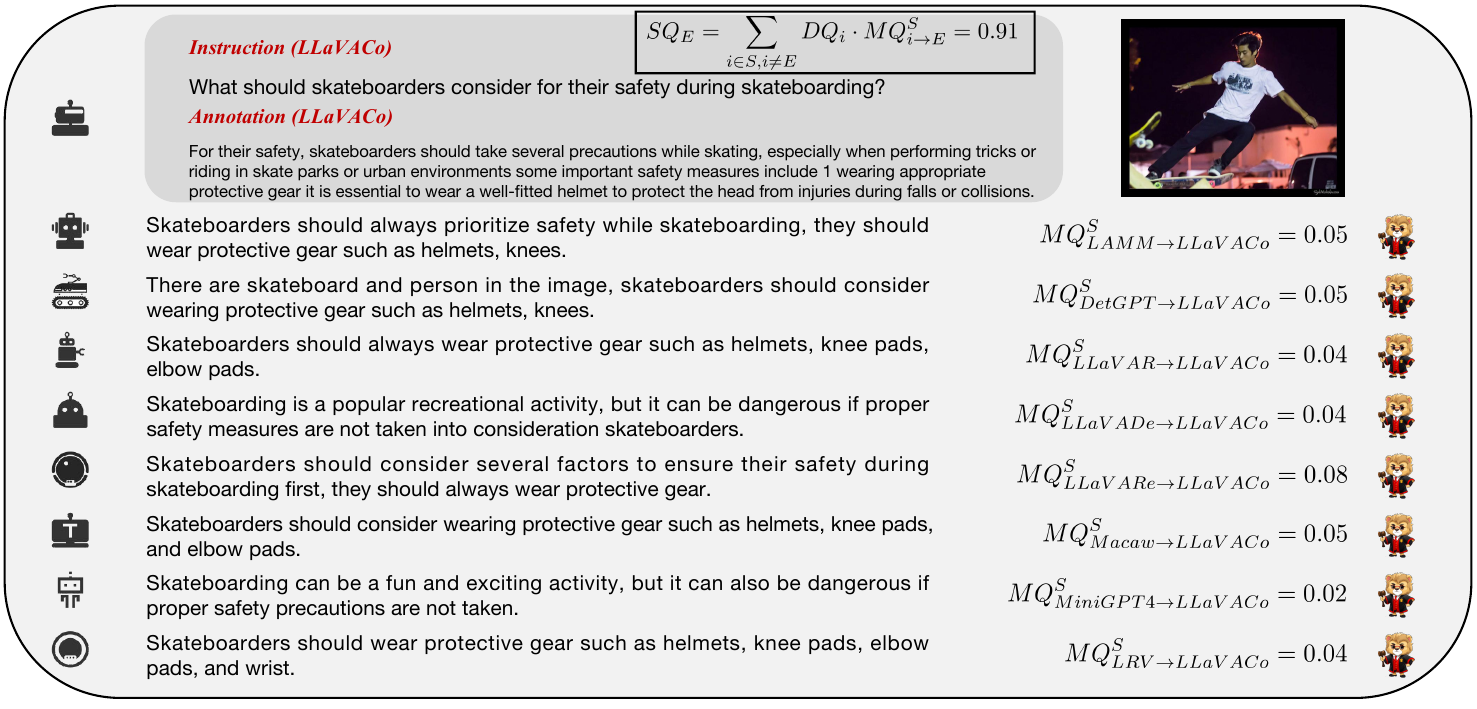}}
  \centerline{}
\end{minipage}
\caption{A conversation sample in LLaVA-Conversation with low SQ measured by other datasets.} 
\label{fig:LLaVACo_low}
\end{figure*}

\begin{figure*}[tb!]
\centering
\includegraphics[width=12.6cm]{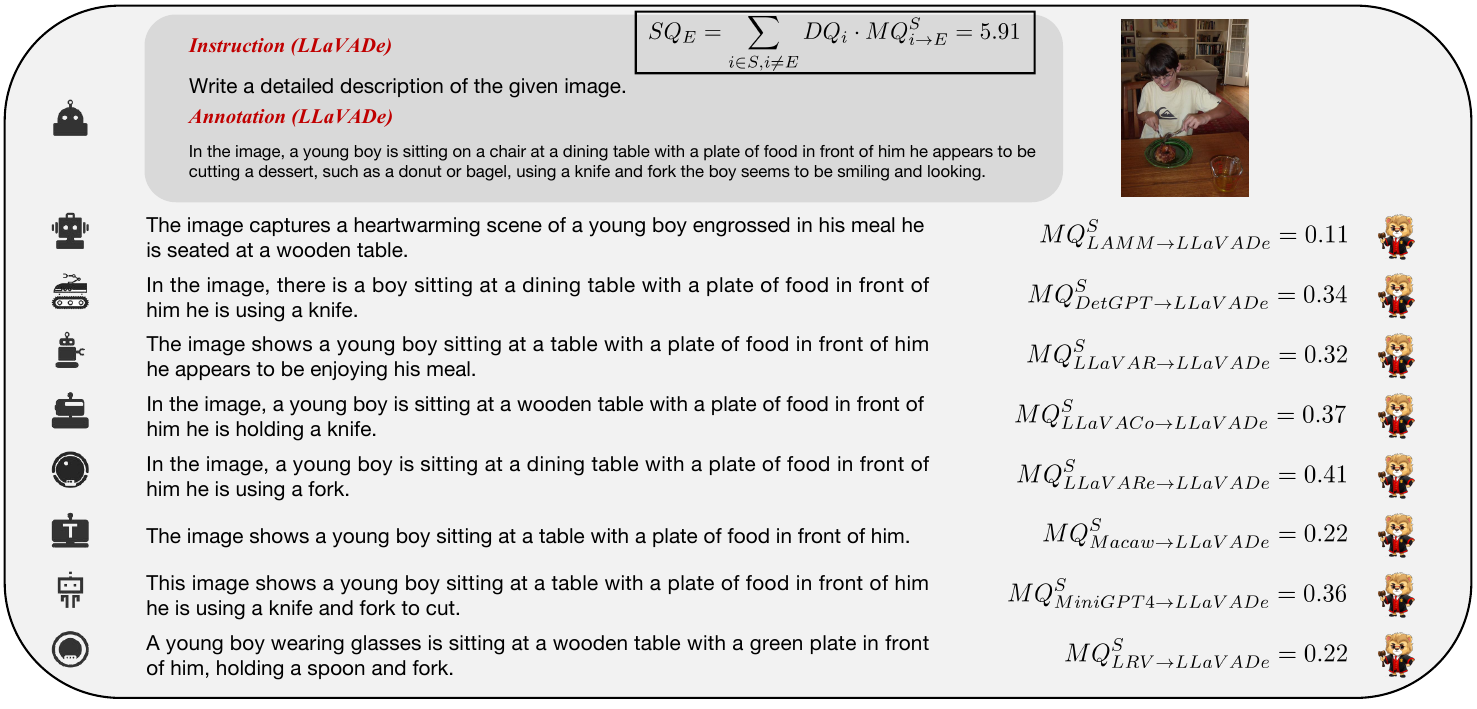}
\caption{A sample in LLaVA-Detailed description with high SQ measured by other datasets.} 
\label{fig:LLaVADe_high}
\end{figure*}

\begin{figure*}[tb!]
\centering
\includegraphics[width=12.6cm]{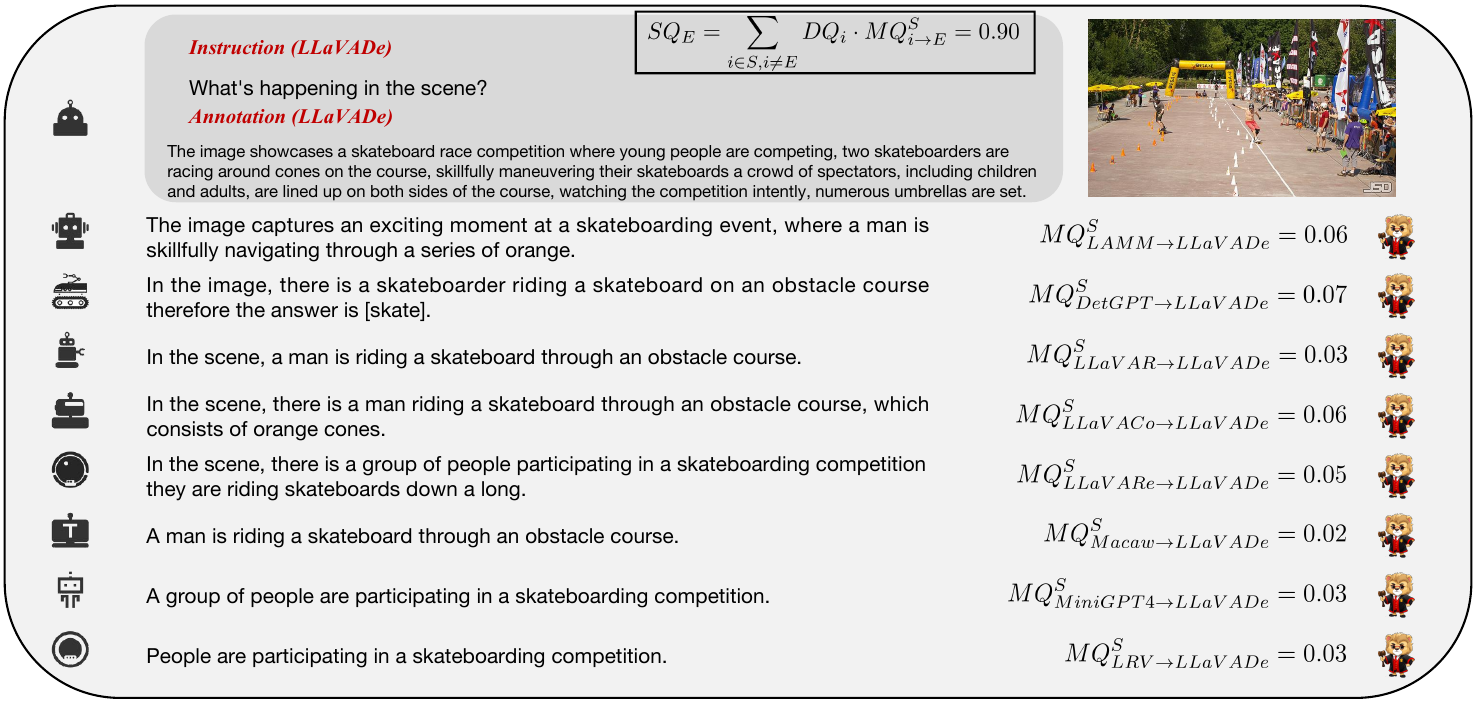}
\caption{A sample in LLaVA-Detailed description with low SQ measured by other datasets.} 
\label{fig:LLaVADe_low}
\end{figure*}

\begin{figure*}[tb!]
\centering
\includegraphics[width=12.6cm]{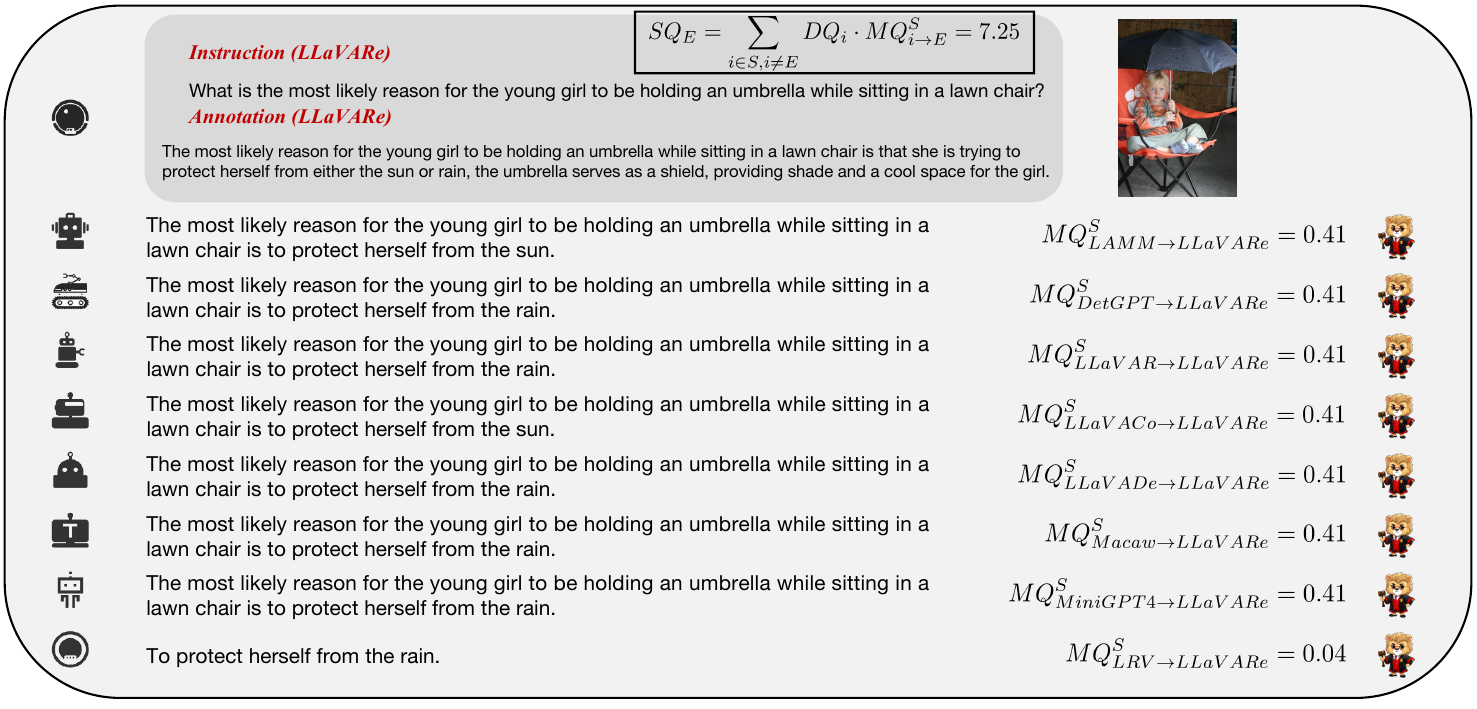}
\caption{A sample in LLaVA-Reasoning with high SQ measured by other datasets.} 
\label{fig:LLaVARe_high}
\end{figure*}

\begin{figure*}[tb!]
\centering
\includegraphics[width=12.6cm]{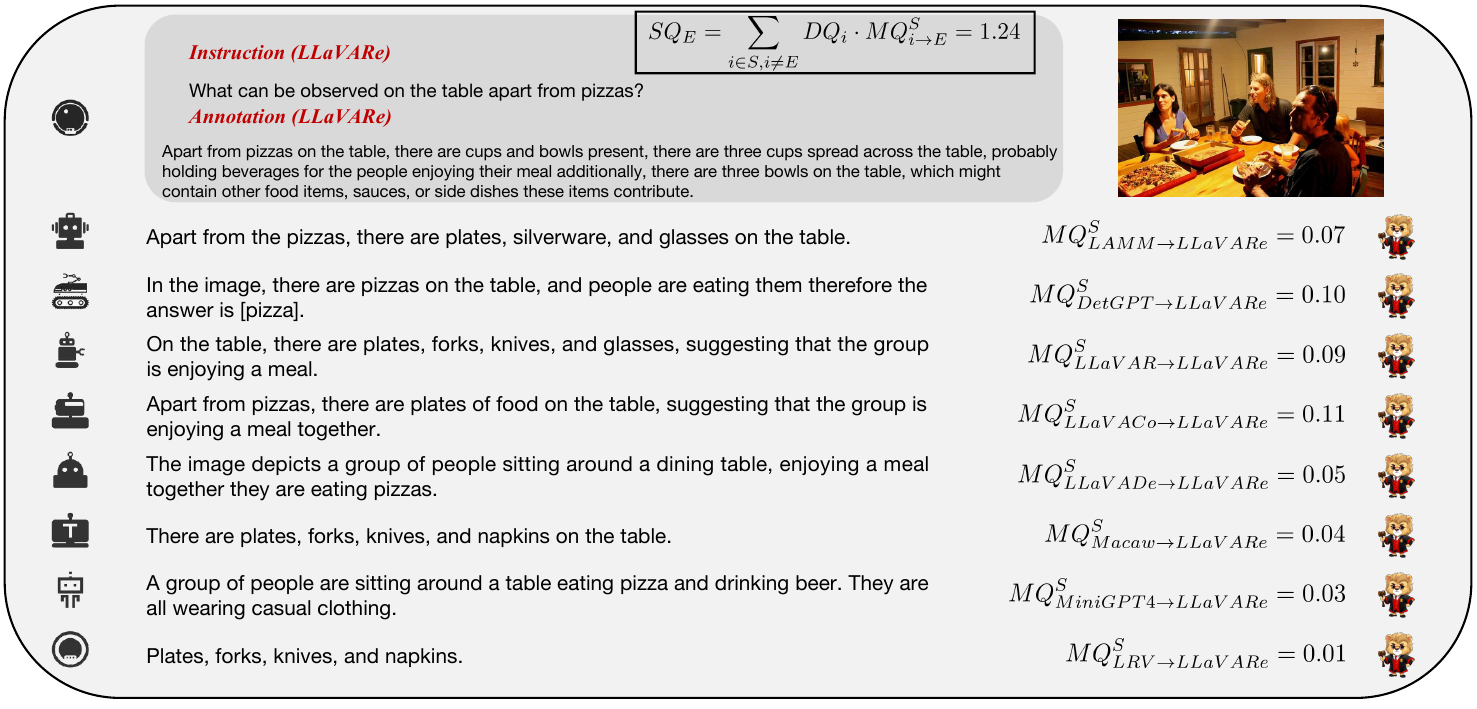}
\caption{A sample in LLaVA-Reasoning with low SQ measured by other datasets.} 
\label{fig:LLaVARe_low}
\end{figure*}

\begin{figure*}[tb!]
\centering
\includegraphics[width=12.6cm]{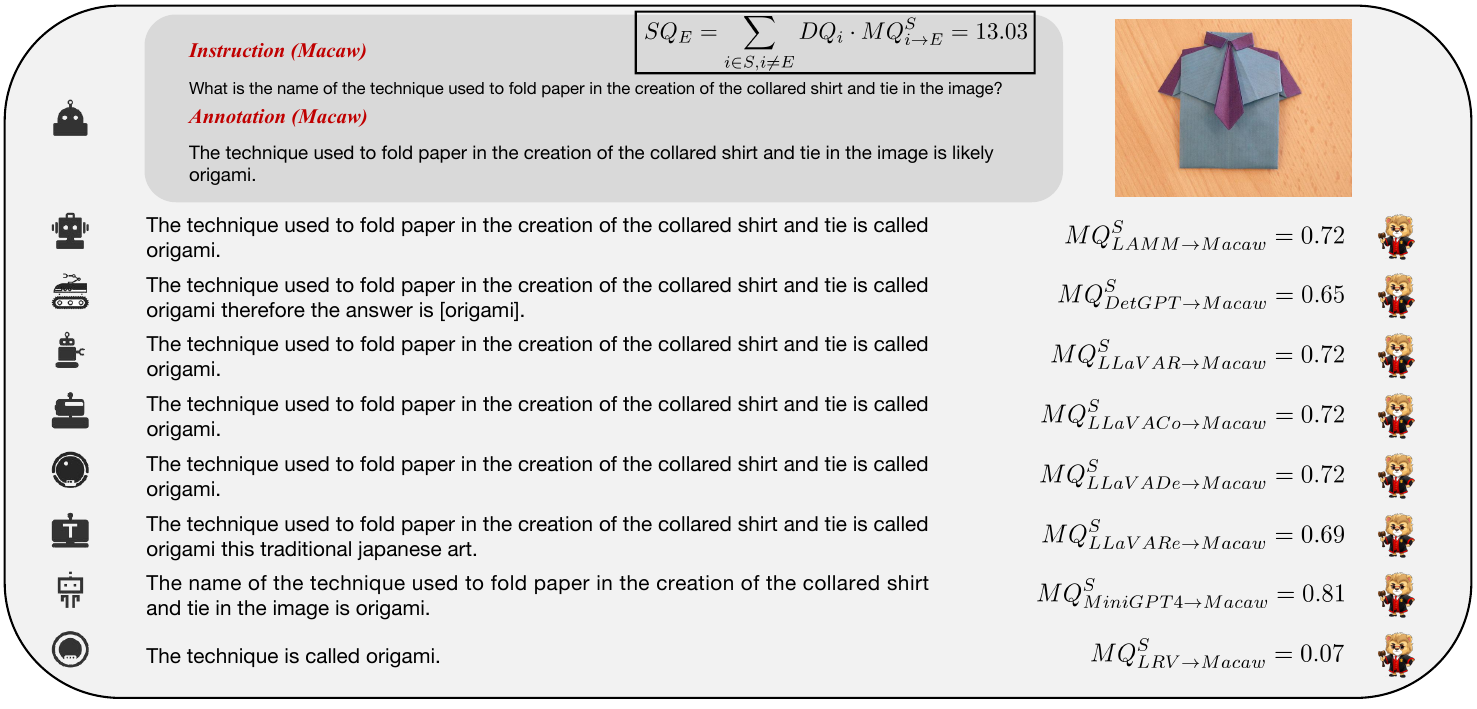}
\caption{A sample in Macaw with high SQ measured by other datasets.} 
\label{fig:Macaw_high}
\end{figure*}

\begin{figure*}[tb!]
\centering
\includegraphics[width=12.6cm]{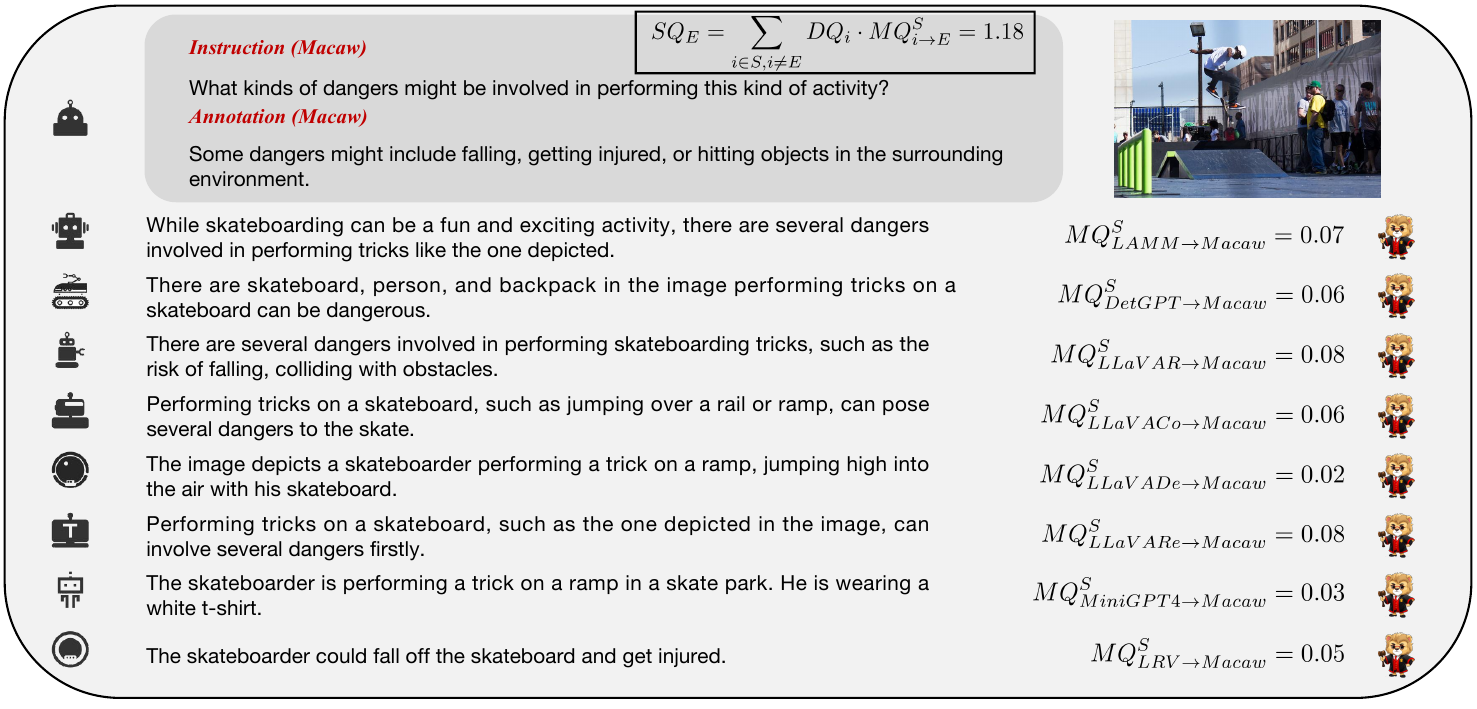}
\caption{A sample in Macaw with low SQ measured by other datasets.} 
\label{fig:Macaw_low}
\end{figure*}

\begin{figure*}[tb!]
\centering
\includegraphics[width=12.6cm]{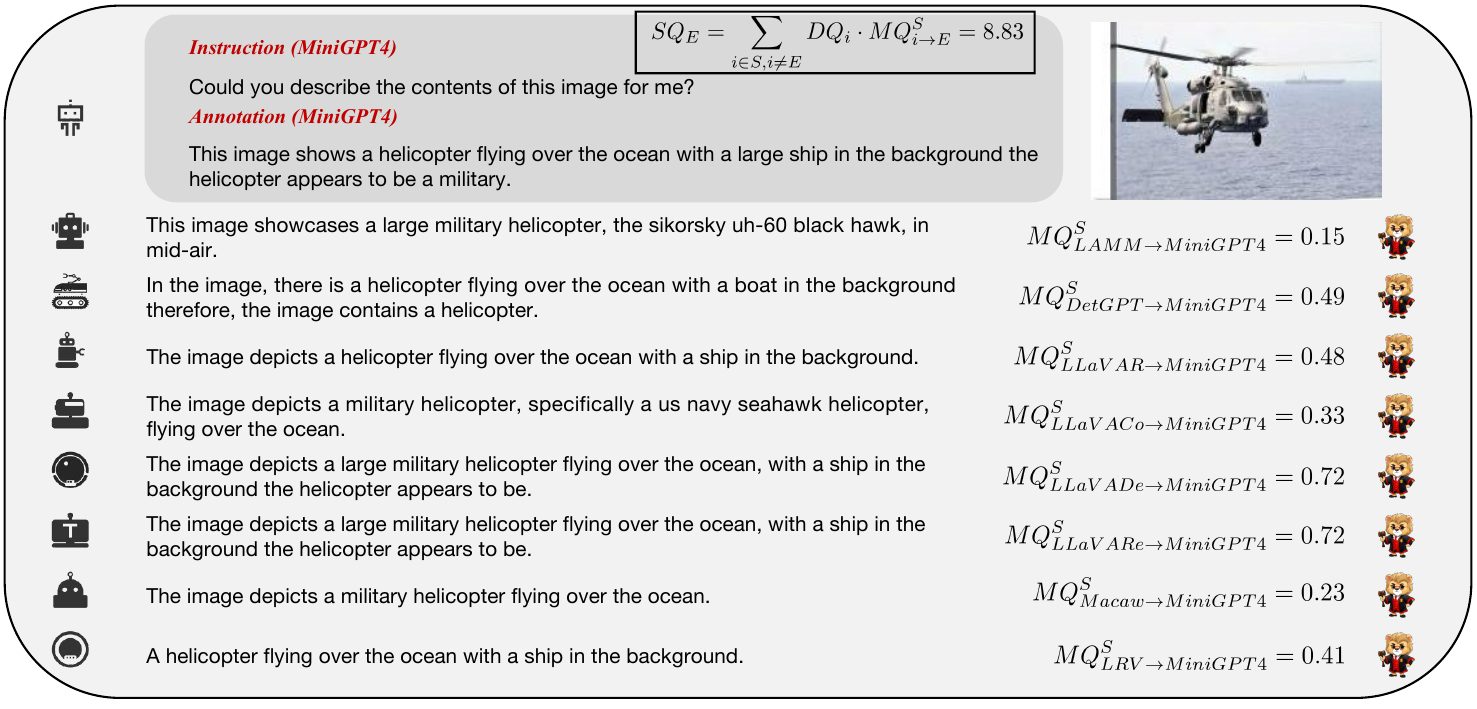}
\caption{A sample in MiniGPT-4 with high SQ measured by other datasets.} 
\label{fig:MiniGPT4_high}
\end{figure*}

\begin{figure*}[tb!]
\centering
\includegraphics[width=12.6cm]{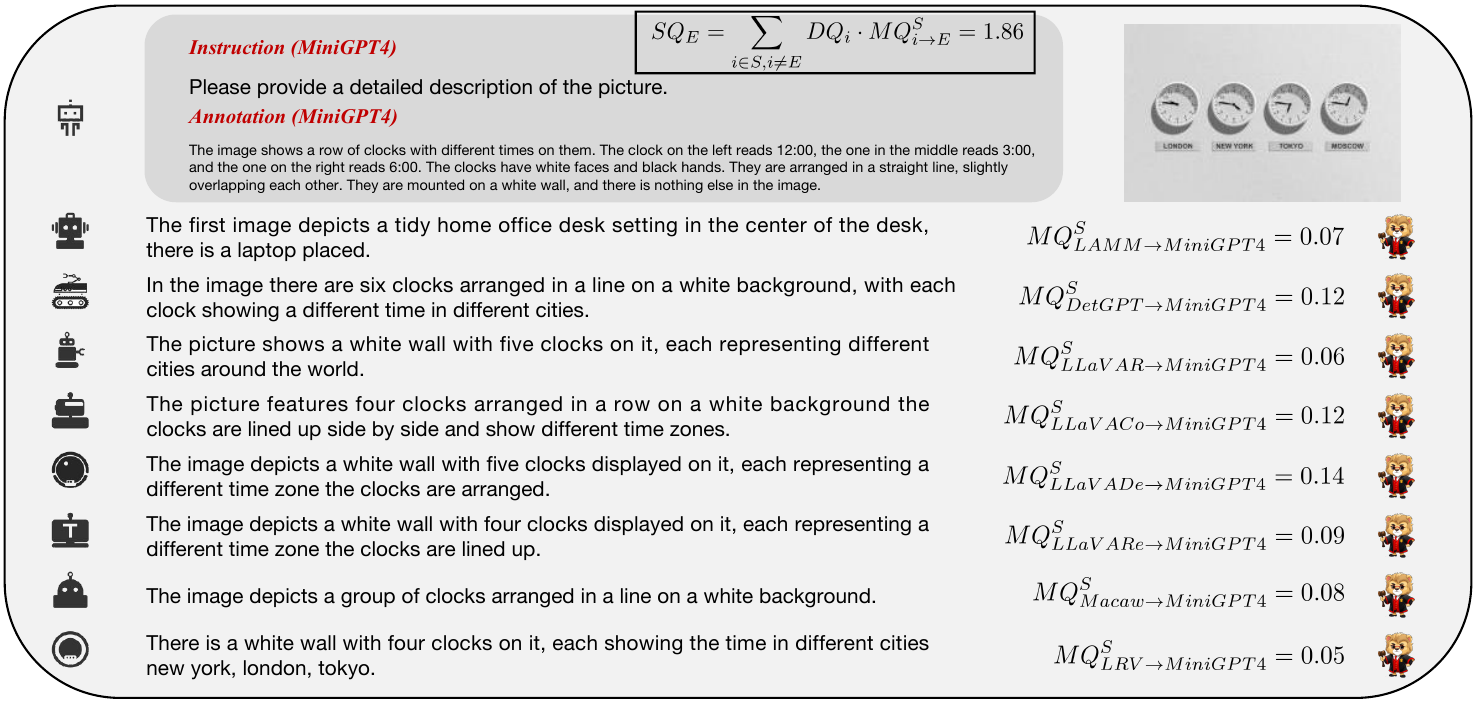}
\caption{A sample in MiniGPT-4 with low SQ measured by other datasets.} 
\label{fig:MiniGPT4_low}
\end{figure*}

\begin{figure*}[tb!]
\begin{minipage}[b]{16cm}
  \centering
  \centerline{\includegraphics[width=12.6cm]{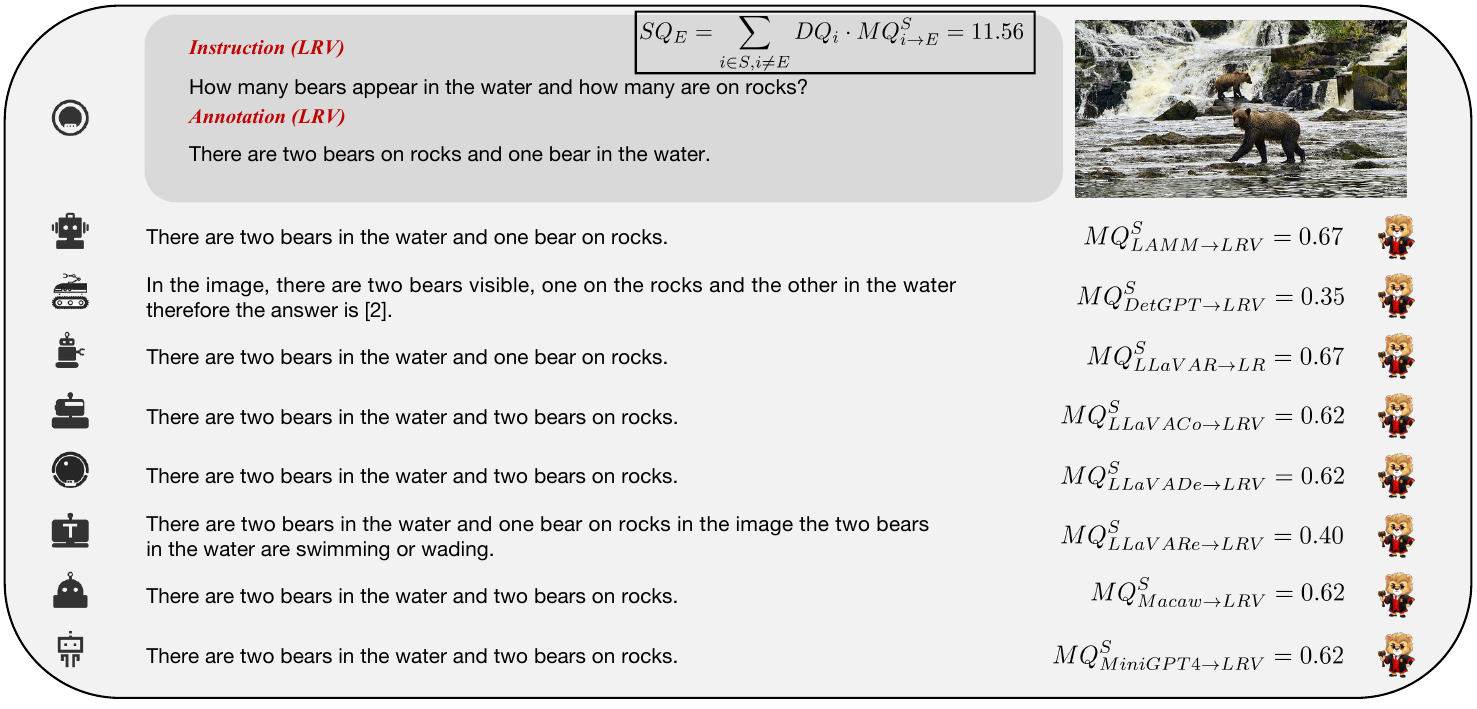}}
  \centerline{}
\end{minipage}

\begin{minipage}[b]{16cm}
  \centering
  \centerline{\includegraphics[width=12.6cm]{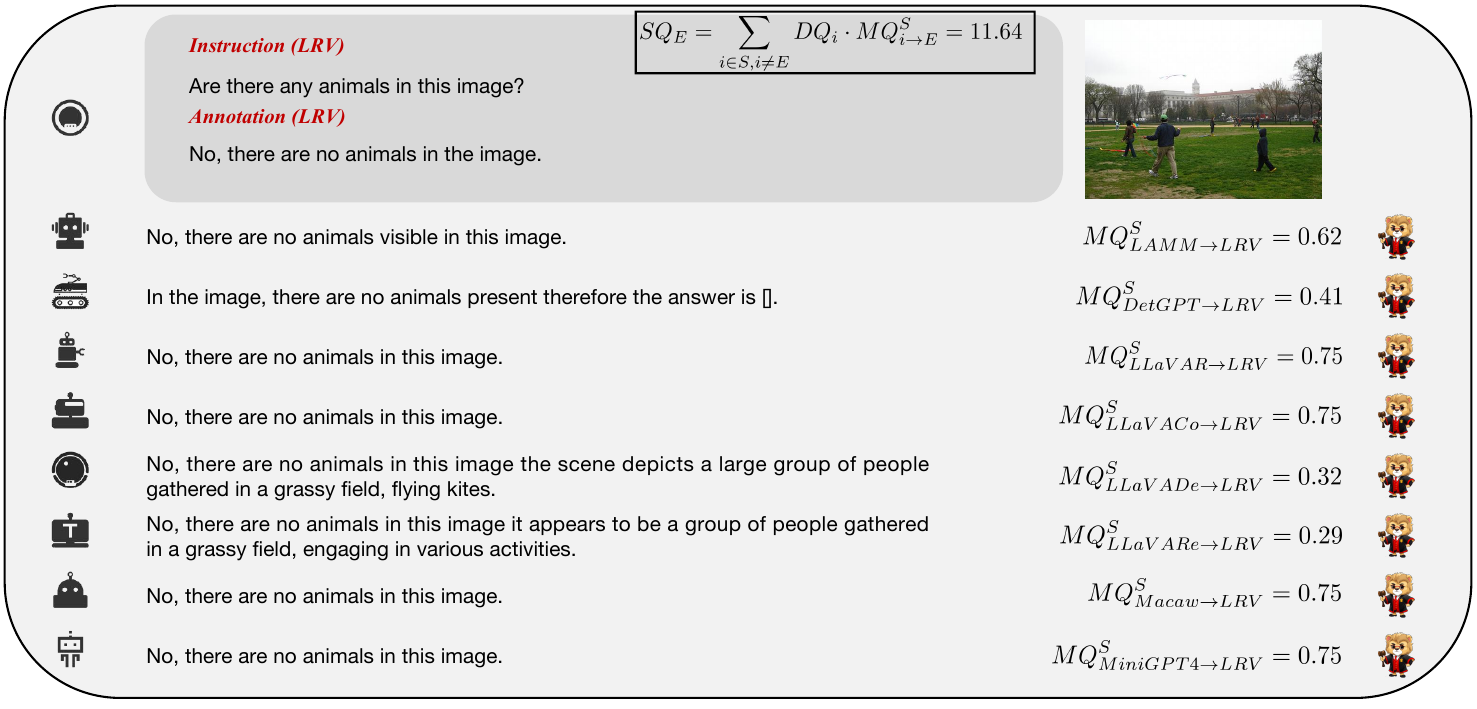}}
  \centerline{}
\end{minipage}
\caption{Two samples in LRV with high SQ measured by other datasets.} 
\label{fig:LRV_high}
\end{figure*}

\begin{figure*}[tb!]
\centering
\includegraphics[width=12.6cm]{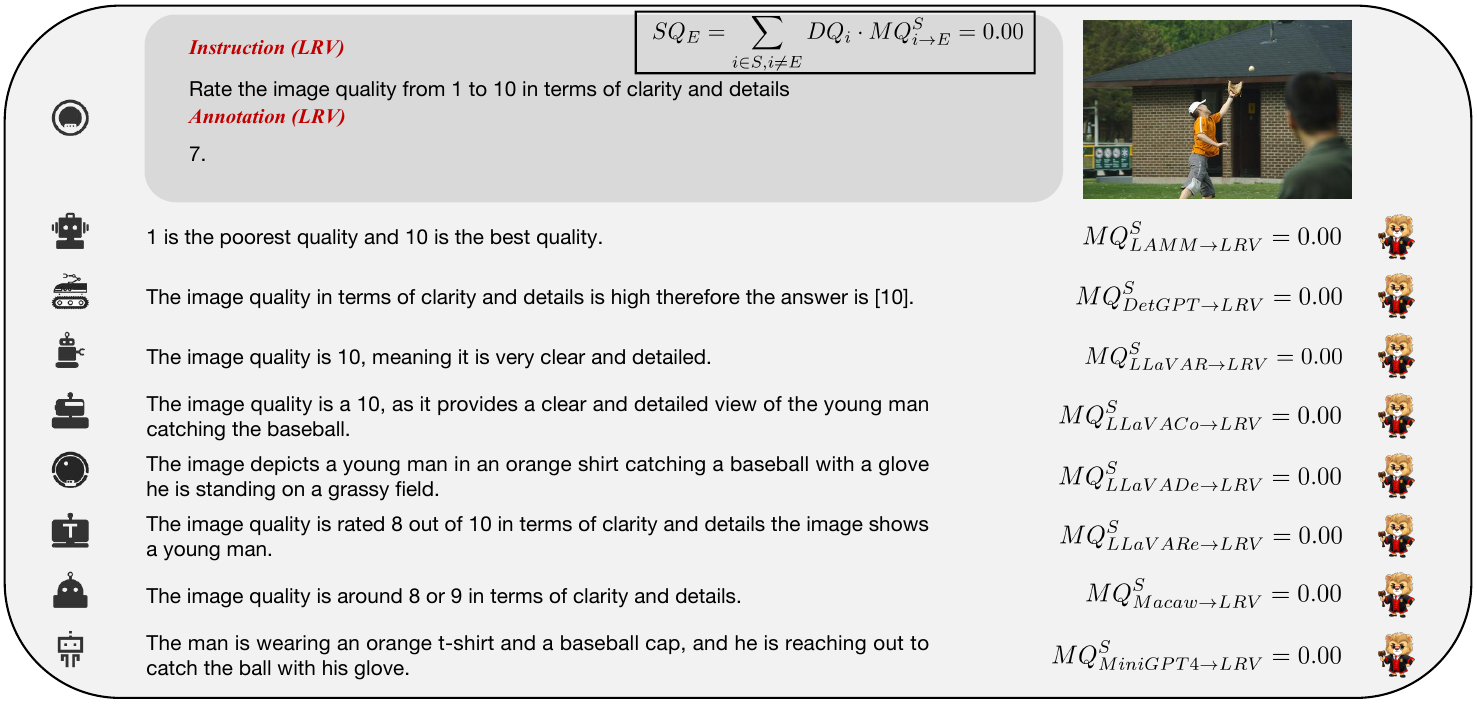}
\caption{A sample in LRV with low SQ measured by other datasets.} 
\label{fig:LRV_low}
\end{figure*}

Based on the proposed \textit{tune-cross-evaluation} paradigm, for each dataset, we show several samples with high SQ and low SQ from Fig.~\ref{fig:DetGPT_low} to Fig.~\ref{fig:LRV_low}.

\end{document}